%% file: main.tex
\documentclass[11pt, a4paper, logo]{zeus}

\usepackage[utf8]{inputenc} % allow utf-8 input
\usepackage[T1]{fontenc}    % use 8-bit T1 fonts
\usepackage{hyperref}
\hypersetup{colorlinks,allcolors=black}
\usepackage{url}            % simple URL typesetting
\usepackage{booktabs}       % professional-quality tables
\usepackage{amsfonts}       % blackboard math symbols
\usepackage{nicefrac}       % compact symbols for 1/2, etc.
\usepackage{microtype}      % microtypography
\usepackage{lipsum}
\usepackage{fancyhdr}       % header
\usepackage{graphicx}       % graphics
\graphicspath{{media/}}     % organize your images and other figures under media/ folder
\usepackage{tabularx}
\usepackage{amsmath}
\usepackage{xspace}
\usepackage{subcaption}
\usepackage{booktabs}
\usepackage{siunitx}
\usepackage{xcolor}
\usepackage{multirow}

\usepackage{titletoc}
\usepackage{tikz}
\usetikzlibrary{patterns}
\usepackage[sorting=none,defernumbers=false]{biblatex}

\usepackage[most]{tcolorbox}

\definecolor{geoGreen}{RGB}{52, 156, 85}
\definecolor{leoBlue}{RGB}{22, 104, 189}
\definecolor{staticOrange}{RGB}{255, 179, 24}

\newcommand{\GEOCircle}[1][0.9]{%
    \tikz[baseline=(char.base)]\node[shape=circle,draw=black,inner sep=2pt,line width=1pt,fill=geoGreen,text=white,scale=#1] (char) {G};\hspace{-1pt}
}

\newcommand{\LEOCircle}[1][0.9]{%
    \tikz[baseline=(char.base)]\node[shape=circle,draw=black,inner sep=2pt,line width=1pt,fill=leoBlue,text=white,scale=#1] (char) {L};\hspace{-1pt}
}

\newcommand{\STATICCircle}[1][0.9]{%
    \tikz[baseline=(char.base)]\node[shape=circle,draw=black,inner sep=2pt,line width=1pt,fill=staticOrange,text=white,scale=#1] (char) {S};\hspace{-1pt}
}

\newcommand{\ourmodel}{EarthNet\xspace}

\title{Global atmospheric data assimilation with multi-modal masked autoencoders}

\author[1]{Thomas J. Vandal}
\author[1]{Kate Duffy}
\author[1]{Daniel McDuff}
\author[1]{Yoni Nachmany}
\author[1]{Chris Hartshorn}
\affil[1]{Zeus AI}

\correspondingauthor{tj@myzeus.ai}

\keywords{Data assimilation, multi-modal foundation models, microwave sounders, infrared imagers, weather forecasting}

\date{July 2024}

\begin{document}

\begin{refsection}

\begin{abstract}
Global data assimilation enables weather forecasting at all scales and provides valuable data for studying the Earth system.
% Operational systems for global data assimilation ingest weather observations using computationally expensive physics based operations that limit the effectiveness of diverse observations. 
However, the computational demands of physics-based algorithms used in operational systems limits the volume and diversity of observations that are assimilated.
Here, we present ``\ourmodel'', a multi-modal foundation model for data assimilation that learns to predict a global gap-filled atmospheric state solely from satellite observations.
% from satellite and other types of weather observations without using any reanalysis datasets. 
\ourmodel is trained as a masked autoencoder that ingests a 12 hour sequence of observations and learns to fill missing data from other sensors.
% We show that \ourmodel is capable of reconstructing missing sensors and predicting global 0.16$^\circ$ analysis data of 3D atmospheric temperature and humidity. 
We show that \ourmodel performs a form of data assimilation producing a global 0.16$^\circ$ reanalysis dataset of 3D atmospheric temperature and humidity at a fraction of the time compared to operational systems. 
It is shown that the resulting reanalysis dataset reproduces climatology by evaluating a 1 hour forecast background state against observations.
We also show that our 3D humidity predictions outperform MERRA-2 and ERA5 reanalyses by 10\% to 60\% 
between the middle troposphere and lower stratosphere (5 to 20 km altitude) 
% (between 50 and 500 hPa)
and our 3D temperature and humidity are statistically equivalent to the
Microwave integrated Retrieval System (MiRS) observations at nearly every level of the atmosphere.
% \ourmodel is shown to represent extreme events such convective potential and heat waves.
Our results indicate significant promise in using \ourmodel for high-frequency data assimilation and global weather forecasting. 
\end{abstract}

\maketitle

% Figures
% 1. Any-to-any architecture with different sensors
% 2. Masked training samples and reconstruction
% 3. Global reconstructions of modalities
% 4. Performance versus operational systems
% 5. 3D vertical structure reconstruction against radiosondes 

\section*{Introduction}

Observations of the Earth's atmosphere, land, and ocean are invaluable for studying the planet and predicting future conditions.  
These observations are ingested by data assimilation (DA) systems operated on some of the world's largest supercomputers by organizations like National Aeronautics and Space Administration (NASA), National Oceanic and Atmospheric Administration (NOAA), and the European Center for Medium-Range Weather Forecasting (ECMWF).
DA in numerical weather prediction (NWP) merges observations with forecasts through statistical methods to produce an optimal initial state of the atmosphere.
The quantity, quality, and diversity of observations is accelerating as a result of combined public and private sector investments and technological advancements \cite{joppa2017case}. 
While these data improvements provide us with more overall information, most is never used to inform weather forecasts. 

Operational DA systems ingest a wide variety of observations to produce analysis ready data and to initialize NWP. 
These petabyte/exabyte scale data processing steps translate raw observations from satellites into geophysical variables that are then ingested by variational DA methods.  
Every six hours, DA systems at NOAA \cite{rodell2004global} and ECMWF \cite{hersbach2020era5} collect the most recent observations and perform this process to generate new forecasts.
The process takes 3-5 hours to complete with the majority of the computational cost from applying 3D/4D variational DA. 
Weather forecasts are then distributed many hours after the observations occurred, creating a large gap between reality and forecast. 
Forecast data produced by NOAA and ECMWF are used around the world for decision making and emergency response. 

In recent years, numerous studies have shown impressive results replacing physics-based numerical weather prediction with AI models, both in terms of improved computational efficiency \emph{and} accuracy \cite{bauer2015quiet,lam2023learning,bi2023accurate,zhang2023skilful,ebert2023outlook}.
Improvements are found by learning model dynamics from reanalysis data that are not well represented in the physics-based NWP models. 
The efficiency and accuracy gains for short-term forecasting indicate substantial potential for long-term adoption of the technology. 
However, current operational AI forecasts depend on global DA systems that are not learned from data.
This presents a large gap between observations and the weather forecasts released to users.

A more efficient DA system that learns from observations and can be applied in near real-time would generate more accurate and higher resolution representations of the environment, leading to improved forecasts.
As highlighted in the National Academy's Decadal Survey \cite{national2007earth,board2019thriving}, modeling the 3D atmospheric structure of temperature, humidity, and winds is a key challenge in current NWP systems. 
A solution would be ``transformative to weather and air quality forecasting'' \cite{board2019thriving}.
Improving atmospheric profile accuracy and resolution is important for both global scale energy balance and local short-term severe events.
The key challenge is obtaining complete vertical profile observations, globally, in all-sky conditions, at resolutions applicable to forecasting and representing these well in a geophysical model.
Here we present an approach to ingest general weather observations aimed at producing more timely and accurate 3D profiles of temperature and humidity using a pure AI approach.

\begin{center}
\begin{table}[htbp]
\small
\centering
\begin{tabular}{lll} 
\toprule[1.5pt]
\textbf{Modality / Sensor}  & \textbf{Orbit} &  \textbf{Channels / Variables}  \\
 \hline
GOES-16 Advanced Baseline Imager (ABI) \cite{schmit2017closer} & \GEOCircle & Thermal infrared 10 bands\\
GOES-18 Advanced Baseline Imager (ABI) \cite{schmit2017closer} & \GEOCircle & Thermal infrared 10 bands \\
Geostationary Korea Multi-Purpose Satellite - 2A (GK2A) \cite{chung2020meteorological} & \GEOCircle & Thermal infrared 10 bands \\
Spinning Enhanced Visible Infra-Red Imager (SEVIRI) \cite{aminou2002msg} & \GEOCircle & Thermal infrared 8 bands \\
Advanced Technology Microwave Sounder (ATMS) \cite{weng2012introduction} & \LEOCircle & Brightness temperature 22 bands \\
Visible Infrared Imaging Radiometer Suite (VIIRS) \cite{murphy2006visible} & \LEOCircle & Thermal infrared 7 bands \\ 
\hline
Shuttle Radar Topography Mission (SRTM) \cite{farr2007shuttle} & \STATICCircle & Elevation, land-sea mask \\
\hline
Microwave integrated Retrieval System temperature \cite{boukabara2011mirs} & \LEOCircle & 3D temperature (37 levels) \\
Microwave integrated Retrieval System humidity \cite{boukabara2011mirs} & \LEOCircle & 3D specific humidity (37 levels) \\
%Microwave infrared Retrieval System (MiRS) & LEO & Skin temperature, surface pressure, emissivity,  \\
%'TSkin', 'SurfP', 'Emis', 'TPW', 'RWP', 'CLW'
\bottomrule[1.5pt]
\end{tabular}
\caption{\small\textbf{Sensor modalities assimilated in \ourmodel{}} comprise a diversity of spectra and orbital perspectives. Along with static topographical data, these sources provide complementary views of atmospheric and surface states and enable the model to learn complex relationships across space, time, and modality. \\ \GEOCircle=Geostationary (GEO), \LEOCircle=Low Earth Orbit (LEO), \STATICCircle=Static.}
\label{tab:sensors}
\end{table}
\end{center}

\begin{figure}[t]
    \centering
    \includegraphics[width=0.95\linewidth]{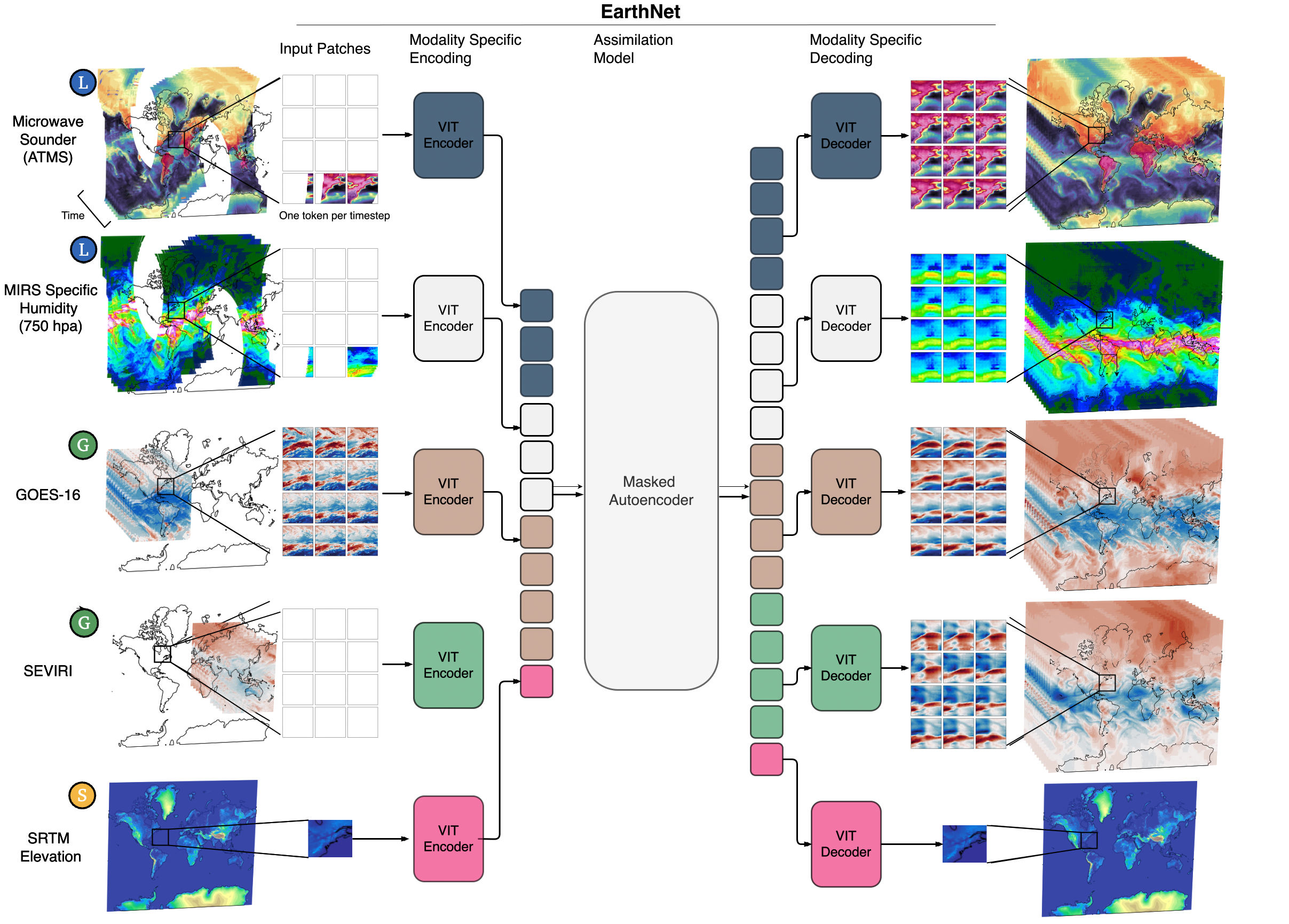}
    \caption{\small{\textbf{\ourmodel ingests multi-dimensional Earth observations} from varying orbits and spectra. Sensor modalities in the first four rows have 12 hours of input sequence with a number of channels. The last row is a static elevation variable defining the topography. Sub-images are extracted spatially of size $(144, 144)$ with a token size of $(16, 16)$. Tokens are encoded with a vision transformer and embedded per sensor modality. After tokens pass through the backbone transformer, each decoder sees all context tokens. This process is applied as a moving window across the image and reassembled using Hann windows.}}
    \label{fig1:overview}
\end{figure}

\section*{\ourmodel}

We propose Multi-modal Masked Autoencoders (MMAE) as an alternative approach to DA that includes end-to-end learning of observational and forward operators from historical observations.
Masked Autoencoders (MAE) are widely used in large language and vision models to learn expressive representations applicable to downstream tasks \cite{he2022masked}.
MAEs define the learning problem such that a subset of input features predicts the remaining features, typically using a transformer architecture and trained to maximize the model's marginal likelihood \cite{moreno2023masked, reed2023scale}. 
Applied to Earth observations, MAEs can be used to gap-fill areas with sparse or missing observations if given sufficient context from different sensors and nearby in space-time. 
% In the vision context, it has been found that MAEs can gap-fill 75\%/90\% of images/videos pixels \cite{feichtenhofer2022masked}. 
Recent work on MMAE presents a scalable encoder-decoder transformer methodology applicable to diverse modalities with steerable content generation \cite{bachmann2022multimae,mizrahi20244m}. 
Here we extend the MMAE to Earth observations with temporal sequences, multi-spectral imagery, and data gaps.

A key feature of our approach is the strict dependence on learning directly from observations, either from space or ground systems, without any reanalysis datasets.
Nine datasets listed in Table \ref{tab:sensors} are used to train the model including geostationary (GEO), low earth orbit satellite (LEO) observations, and static topographical data.
Data is interpolated and reprojected to hourly, 0.16$^\circ$ resolution composites which is higher resolution than current operational analysis products. 
% Four geostationary datasets with infrared sensors including GOES-16/18 ABI, MSG SEVIRI, and GeoKompSat-2A AMI create a ring around the Earth at $\pm$60$^\circ$ providing constant temporal coverage.
% VIIRS on JPSS-1 is used as infrared data in the polar regions missed by geostationary. 
% Two microwave ATMS sounding datasets from SNPP and JPSS-1 are ingested see through the clouds, extract vertical atmospheric information, and all-sky land surface temperature.
% Lastly, elevation and land-sea mask are taken from the SRTM dataset providing the model context of surface and atmospheric dynamics.
LEO sensors observe every point on Earth approximately twice per day, giving us four ATMS and two VIIRS samples, while GEO is available every hour. 
A composited dataset of two years is collected and transformed from  $\sim$500 terabytes to 2 terabytes of clean training data.
Data volume is reduced as we downsample sensors to the lowest common spatial resolution of ATMS near 0.16$^\circ$. 
Each modality is a high-dimensional tensor with space, time, and channel dimensions, much like a video.
The combination of four geostationary sensors covers 60\% of the pixels at all times.
Each LEO sensor captures approximately 10\% of the pixels, with 20\% from ATMS soundings.
Given the range of modalities, these ratios are well within the capabilities of masked autoencoders. 
% A training sample includes each modality across a 12 hour window with varying channel counts producing 3D and 4D input tensors.

% https://app.diagrams.net/#G1Kw4yBwRdvv46CrpItdAt-YoN6kD1YEoN#%7B%22pageId%22%3A%22bQT3zpDmAVFkPRLbGovw%22%7D
\begin{figure}[t]
    \centering
    \includegraphics[width=0.95\linewidth]{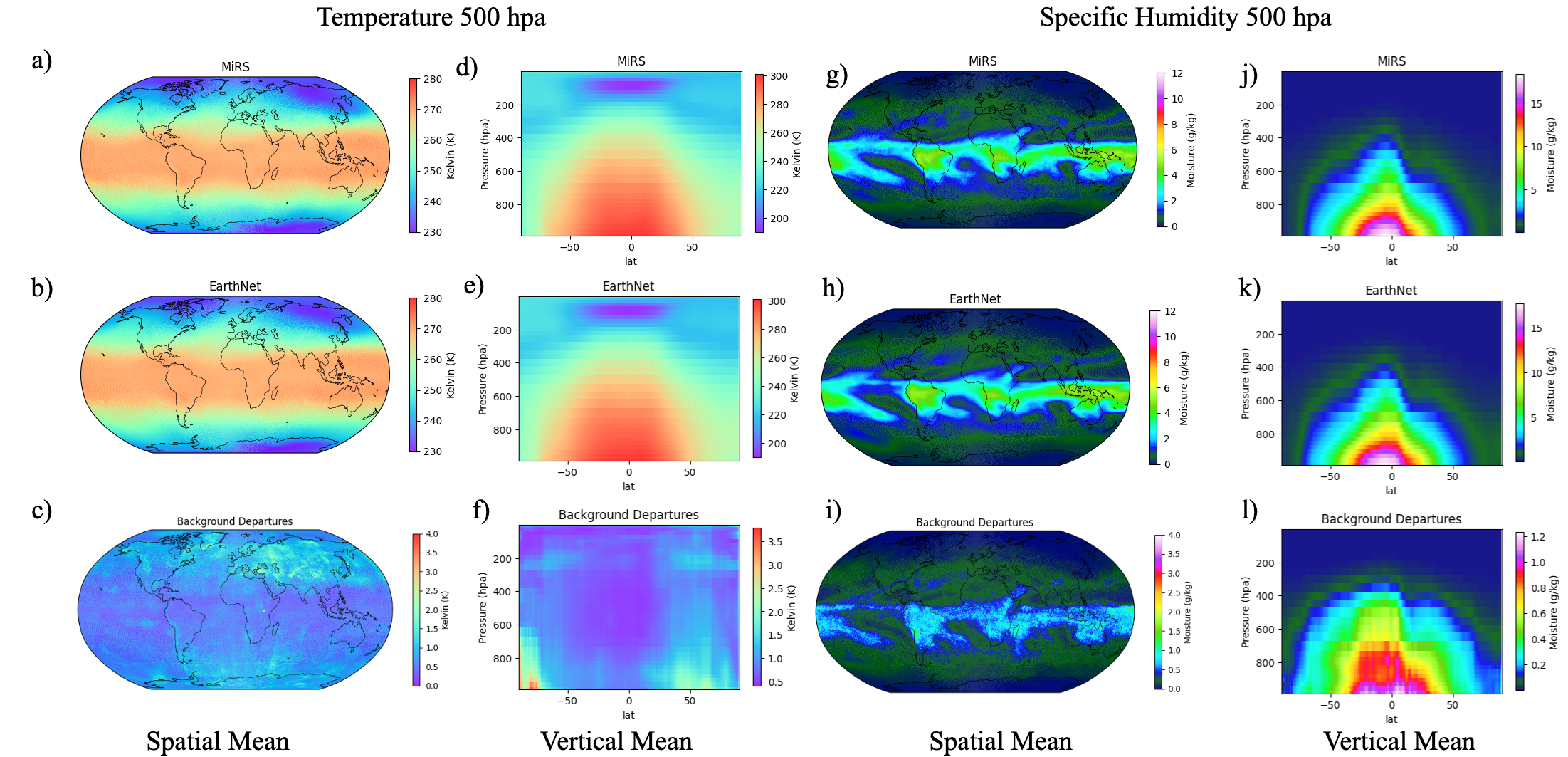}
    \caption{\small{\textbf{Background departures} of \ourmodel's temperature (a-f) and specific humidity (g-l) against observations temporally averaged across the spatial and vertical dimensions.
    The top row shows MiRS average temperature and humidity values across February and March 2024. 
    \ourmodel's 1 hour background state average temperature and humidity are shown in the second row.
    The third row shows error departures as computed from 1 hour background state predictions minus the MiRS observation.}}
    \label{fig:background}
\end{figure}

\ourmodel, outlined in Figure \ref{fig1:overview}, treats each sensor as a separate modality with a unified transformer architecture and modality specific encoder/decoders as applied in \cite{bachmann2022multimae}.
Modalities with a time dimension are tokenized frame by frame with vision transformers, enabling the model to ignore patches with missing data.
% Tokenizers are pretrained individually as variational autoencoders (VAEs) which was found to greatly improve training efficiency.
% Sinusoidal positional embeddings are used for each modality to generate common space-time representations.
Tokens from each modality are concatenated and passed through the multi-modal transformer.
The predicted tokens are used as context to decode complete information from each modality.
Within, each modality decoder, cross-modality embeddings are added to provide context for the output prediction.  
This process makes it possible to generate every sensor from cross sensor contextual information.
It is expected that as more sensors and modalities are added, the context and decoded results will continue to improve.

The model is trained by randomly selecting a small subset (128 or 1.4\%) of tokens, predicting the remainder, and applying a simple mean square error (MSE) loss.
This process naturally ignores areas of missing data and fills the pixels with context from previous timesteps and other satellites.
Model training is performed in stages on 16 Nvidia V100 graphic processor units (GPUs) for $\sim$21 days.
These stages included pretraining tokenizers, training on level 1 sensors, and adding MiRS data with less temporal coverage.
Short-term 1 to 6 hour forecasts are directly available from \ourmodel by filling future timesteps, which will enable us to evaluate the forward process.
The learned dependencies across space, time, and modalities is a powerful feature that will enable seamless integration with additional sensors.

% https://app.diagrams.net/#G1RRtLXt215SuKOYByrWD0tELs1O_8YBD8#%7B%22pageId%22%3A%22ovMqr7hQWwv2VzhcBFRi%22%7D
\begin{figure}
    \centering
    \includegraphics[width=\linewidth]{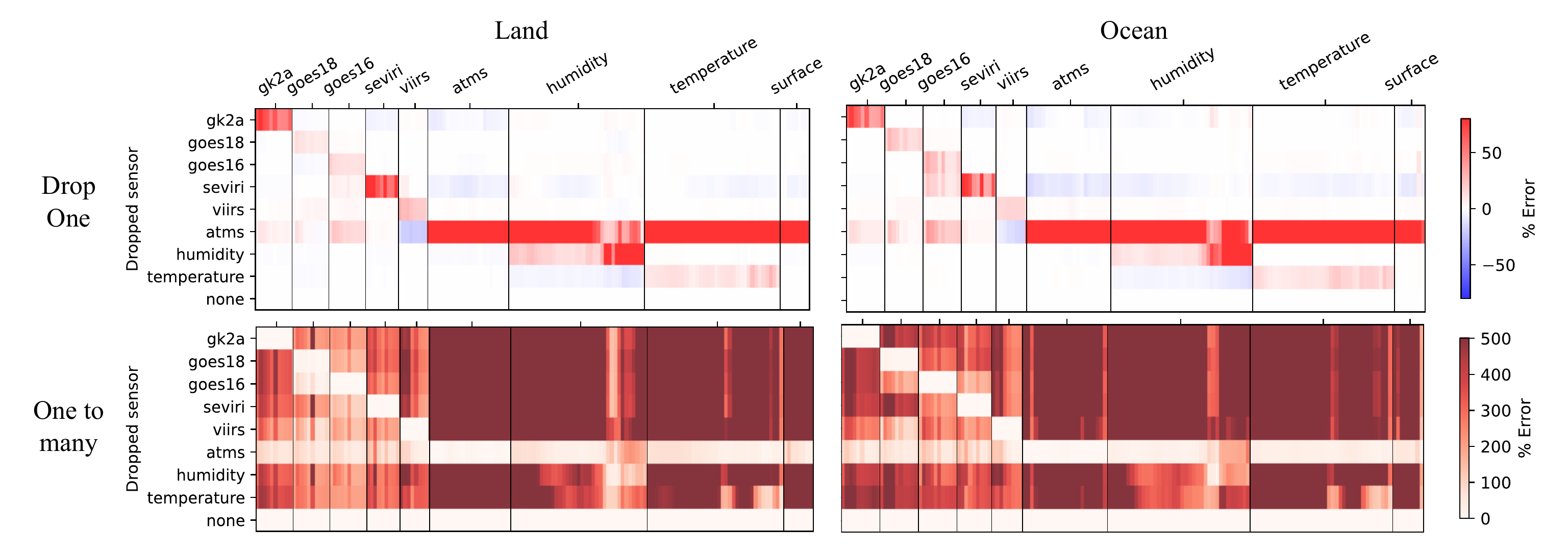}
    \caption{\small{\textbf{Sensitivity analysis and sensor importance} measured by relative mean absolute errors over the land and ocean. (Top row) Each sensor is dropped individually while reconstructing all modalities and comparing errors to the baseline of all modalities included. (Bottom row) One sensor is taken as input to reconstruct all with errors computed as above. The analysis is split into the land (left) and ocean (right) regions to delineate surface types.}}
    \label{fig:sensitivity}
\end{figure}

\section*{Verification Methods}

\ourmodel is evaluated for the ability to reconstruct missing information and generate 3D atmospheric temperature and humidity profiles. 
A test set of February and March 2024 is held out from training and used for each experiment.
We present a series of analyses of the model's capabilities.
First, an evaluation of background state departure errors is presented. 
A sensor sensitivity analysis tests the importance of each modality for reconstruction. 
Comparisons with unseen (by \ourmodel) radiosonde observations against ERA5 \cite{hersbach2020era5} and MERRA-2 \cite{gelaro2017modern} indicate statistically significant global reanalysis skill. 
% Lastly, case studies of two atmospheric events including convective thunderstorms and snowfall shows \ourmodel captures thermodynamic relationships across variables. 

\subsection*{Background departures}
\label{sec:backgrounddepartures}

% Compare 0 step and 1 step reconstrution errors
DA performance is commonly evaluated by comparing the model's background state to observations. 
In traditional DA, this is done by computing the error between a forecast and observations, before those observations are assimilated.
This process increases the independence between inputs and analysis data for a fair evaluation.
Here we perform this experiment by removing the last frame from each input modality then forecasting the last frame.
This is applied for every time step in the test set and saved as the model background state.

Background departures are computed as the mean absolute error (MAE) between the one hour forecast and observations unseen during inference. 
The results presented in Figure \ref{fig:background} show MiRS temperature and humidity 500 hectopascal (hpa) statistics of the spatial and vertical dimensions.
Mean climatologies of MiRS and \ourmodel over the test period are computed and shown in the first two rows.  
It is seen that \ourmodel reconstructs the spatial and vertical distribution of both temperature and humidity effectively. 
Due to the test set months, the highest temperatures are seen near the surface just below the equator. 
This also corresponds to higher humidity values in the tropics.  
Background departures are shown on the bottom row. 
Vertically, temperature errors largely follow the expected distribution with the largest near the surface.

\ourmodel is shown to capture the overall global distribution of specific humidity, both vertically and spatially, though the spatial distribution of humidity is smoother in \ourmodel indicating degraded resolution.
Vertically, \ourmodel well captures the boundary layer height and distribution. 
Background departures are mostly limited to 1 g/kg and are found largest in high humidity tropical regions. 
Overall we see that \ourmodel captures the global spatial and vertical distributions of temperature and humidity. 
Detailed statistics of bias and MAE for all modalities and channels can be found in the Supplement \ref{sup:sec:backgrounddepartures}.

\subsection*{Sensor sensitivity analysis}

By exploring the model's ability to perform many-to-many and one-to-many prediction, we gain insight into the importance of each sensor for representing the atmosphere's state.
Here we perform a sensor sensitivity test by systematically removing one or more sensors to reconstruct all. 
In this way we are able to remove information at inference and test the model's ability to reproduce that observation. 

In the first experiment, we remove one sensor at a time from the input and reconstruct all. 
This process is executed for every sensor and compared to the baseline of no sensors dropped. 
Mean absolute errors of observations minus reconstructions are computed over land and ocean regions. 
Results in Figure \ref{fig:sensitivity} (top row) show that as sensors are removed, the mean absolute error generally increases, indicating that more sensors would improve performance. 
We find that ATMS has a high effect in predicting humidity and temperature while infrared imagery is generally less important.
Errors over the ocean regions are slightly larger than over land, most notably for the infrared data. 
It is also found that the overlapping GOES-16 and GOES-18 satellites lose minimal performance, suggesting duplicate information content.
In contrast, GK2A and SEVIRI have little spatial overlap with other GEO sensors and have larger reconstruction errors.
VIIRS contains substantial overlap with GEO sensors outside of the polar regions and is found to have relatively low error when removed. 
% However, it is still expected the infrared imagery from VIIRs and GEO aids in interpolating between ATMS observations.

Secondly, a one-to-many experiment is performed by predicting all sensors from a single input sensor.
This experiment tests the amount of total information from a given sensor to reconstruct all other sensors and the 3D structure. 
Results in Figure \ref{fig:sensitivity} (bottom row) highlight that ATMS contains a substantial amount of information for reconstructing all modalities, including all of the infrared sensors.
This is found over both land and ocean regions, indicating that much of the information in infrared imagery can be represented by microwave sounders.
Infrared sensors show the capability of reconstructing themselves but have limited skill representing the 3D structure. 
The increased uncertainty in the upper troposphere for ATMS and decreased uncertainty for infrared is attributed to errors in ATMS limb correction. 
Overall, these experiments show that microwave sounders provide important information for modeling the atmosphere.
Generally, however, infrared imagery is of higher resolution than sounders due to the sensing technique and in this study the infrared observations are downsampled from 750m/2km to 16km. 
While the utility of infrared imagery is not as apparent in this analysis, the increased temporal frequency is expected to aid in the temporal interpolation of ATMS.

% https://drive.google.com/file/d/1abfWHH_f5HUXTsv_m_BIRyQleLz-RQlQ/view?usp=drive_link

\begin{figure}
    % \centering
    \includegraphics[width=\linewidth]{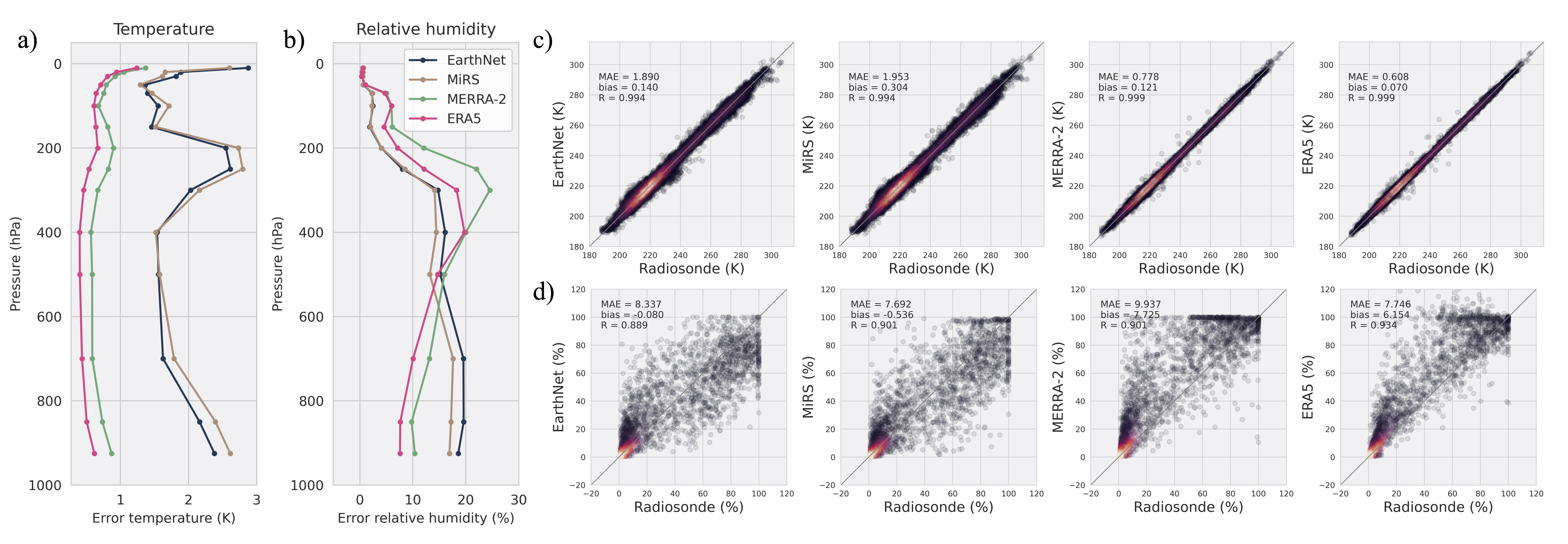}
    \caption{\small{\textbf{Comparison with radiosonde soundings} shows that \ourmodel's performance is similar to MiRS observations in matching radiosonde observations of temperature (a) and humidity (b).
    \ourmodel outperforms MERRA-2 and ERA5 reanalyses for humidity predictions between 50 and 500 hPa,  without the benefit of having assimilated radiosonde data.
    Scatter plots show a strong linear relationship between ERA5/MERRA-2 reanalysis and radiosonde temperature observations (c). 
    Relative humidity scatter plots show more normally distributed errors for  \ourmodel and MiRS at high humidity values (d).
   }}
    \label{fig:reanalysis}
\end{figure}

\subsection*{Global reanalysis verification} 
% figure 2
% this should cover global, spatial, vertical, and statistics
We compared \ourmodel's reanalysis data to those of operational models ERA5 (from ECMWF) and MERRA-2 (from NASA) and Microwave integrated Retrieval System (MiRS) retrievals using radiosonde soundings of temperature and relative humidity as ground truth (Figure \ref{fig:reanalysis}; Table \ref{tab:radiosonde_results}).
Radiosondes are weather balloons designed to directly measure the atmospheric vertical profile at a single location.
We found that \ourmodel errors are similar in magnitude and distribution to MiRS, indicating that \ourmodel is successful at reconstructing MiRS.
Upon inspection of the temperature and relative humidity error distributions over the height of the atmosphere, we find there is no statistically significant difference ($p \le 0.05$) between \ourmodel's errors and MiRS errors at the majority of pressure levels.
Like MiRS, \ourmodel achieves 10-60\% lower error for humidity between 500 and 50 hPa than ERA5 or MERRA-2.
However, we found that \ourmodel has mean absolute error of temperature that is 1.28 K (1.11 K) greater than ERA5 (MERRA-2) over the full vertical distribution.
We note that temperature and humidity errors are relatively consistent between ERA5 and MERRA-2, with ERA5 performing slightly better, perhaps due to the higher spatial resolution of ERA-5 (0.25$^\circ$) compared to MERRA-2 (0.5$^\circ$ x 0.625$^\circ$).

The scatter plots in Figure \ref{fig:reanalysis}c and \ref{fig:reanalysis}d show the correspondence between the reanalysis/satellite products and radiosonde observations.
For temperarure, MERRA-2 and ERA5 show a strong linear fit to radiosonde measurements ($R>0.99$), while
\ourmodel and MiRS have similar and slightly lower correlation ($R=0.94$).
For relative humidity, ERA5 and MERRA-2 exhibit a bias toward higher humidity relative to radiosonde soundings, whereas \ourmodel and MiRS show more normally distributed errors.
\ourmodel's relative humidity has a mean absolute error of 8.337\% that is comparable to numerical models (9.937\% for MERRA-2 and 7.746\% for ERA5) and a minimal bias of $<$1\%, compared to a 6-7\% bias from the numerical models.

These results suggest that \ourmodel is successful at reconstructing MiRS, while highlighting inconsistencies between reference datasets.
Previous studies have noted the unreliability of radiosonde moisture soundings at the 100 hPa level and intravariability between ERA-5, Global Data Assimilation System (GDAS), and MiRS, while concluding that MiRS, and therefore \ourmodel, is ``well within the uncertainty'' of all of the data taken together \cite{boukabara2011mirs}. 
It is also worthwhile to note that radiosonde data are assimilated into DA systems with low uncertainty assignments, potentially drawing ERA5 and MERRA-2 reanalysis disproportionately close to radiosonde observations.

% outperforms operational data assimilation systems ERA5 and MERRA-2 in representing the 3D atmospheric structure of temperature and humidity. 

% \subsection*{Characterizing extreme events}
% Perform nowcasting and test specific events
% Aviation?  Areas with little data? 
% \subsection*{Sounding case studies}
% Convective events in an unstable atmosphere causes damage to society and infrastructure.  Improved representation of convective events in a more timely manner would have large positive consequences. Here we investigate the Convective Available Potential Energy (CAPE) metric to show that \ourmodel ... 

%\subsection*{Extreme temperatures}
%Accurate representation of extreme temperatures, hot and cold, ...

\section*{Discussion}

While incremental NWP improvements have driven slow and steady advancements for decades, recent progress in ML have disrupted this trend and achieved greater accuracy forecasts in some cases than numerical models.
Until now, almost all ML achievements in weather modeling have focused on forecasting and retained a dependence on NWP reanalysis for training and NWP DA for forecast initialization.

We present a versatile model capable of using a masked modeling objective across a range of strictly observational input and output modalities. 
Without any dependence on numerical models, \ourmodel is a multi-modal generative model that can be conditioned on any arbitrary subset of input modalities to produce complete global reanalyses at 0.16$^\circ$.
We show that the model can gap-fill missing observations, perform a one hour forecast, and extend to more modalities.
The methodology outlined has potential to be extended to a wide range of applications, varying resolutions, and modality types.

One of the main limitations of our current architecture are the high memory requirements in both training and inference.
Without further optimization, the high-dimensionality of remote sensing data will limit the number of potential modalities. 
This issue can be addressed by compressing the data into tokens prior to multi-modal training, as shown in \cite{mizrahi20244m}. 
Secondly, due to the encoding transformer token size, the current model does not retain all high-frequency information, leading to overly smooth predictions.

% more limitations
% more use cases

\ourmodel represents a significant departure from traditional DA techniques and showcases the potential of multimodal ML for simulation of Earth system processes.
This advancement arrives as Earth science is experiencing rapid progress in acquiring scientific observations, in a large part due to satellite-based observations that provide a global perspective.
We believe that multi-modal foundation modeling approaches like \ourmodel are poised to make DA possible in near real-time, leading to more timely and accurate short term weather forecasts.

\section*{Acknowledgements}

The computing used to train this model was provided by the NASA Center for Climate Simulation (NCCS) at the Goddard Space Flight Center, accessed through the SBIR award 80NSSC23CA169.
This research also used resources of the National Energy Research
Scientific Computing Center (NERSC), a DOE Office of Science User Facility 
using NERSC award SBIR-ERCAP0030185.
We thank NOAA, NASA, European Space Agency (ESA), and the Korean Meteorology Agency (KMA) for data access. 
Software developed for this study leveraged open source projects including PyTorch and Pangeo.
We thank the thank our collaborators who provided comments on this work: Daniel Holdaway, Tsengdar Lee, Ramakrishna Nemani, and Max Wilson.

%\bibliographystyle{plain}
% \bibliography{references}
\printbibliography[title={References},resetnumbers=true]
\end{refsection}

\newpage 

\begin{refsection}
\include{appendix}
\printbibliography[title={References},resetnumbers=false]
\end{refsection}

\end{document}

%% file: appendix.tex
% \begin{bibunit}

\appendix
\section*{Supplement}

\startcontents[sections]
\printcontents[sections]{}{1}{}

\newpage

%\addcontentsline{toc}{section}{Appendices}
%\renewcommand{\thesubsection}{\Alph{subsection}}

\section{Data}

A range of datasets from satellites to in-situ observations were used for training and verification of \ourmodel.  
Here we outline each dataset used and preprocessing measures.

\subsection{Training data}
\label{appendix:data}

Datasets are all reprojected onto a 0.16$^{\circ}$ global latitude-longitude grid of size (1125, 2249) at a 1 hour resolution.
Observations are rounded to the nearest hour when processed. 

\subsubsection{GOES-16/18}

% Details on GOES-R data specs
Geostationary satellites are synchronized in orbit with Earth's spin to hover over a single location. 
Given this location, the sensor, measuring radiation as often as possible, can frequently capture data over a continuous and large region. 
This feature makes geostationary satellites ideal for capturing environmental dynamics. 
The GOES-R series satellites, namely GOES-16/18 (East and West side of the Americas) operated by NASA and NOAA, provide scientists with unprecedented temporal frequency enabling real-time environmental monitoring using the Advanced Baseline Imager (ABI)~\cite{schmit2017closer}. 
ABI retrieves 16 spectral bands in the visible to infrared wavelengths at a 0.5-2 km spatial and 1-10 minute temporal resolution.
We selected the 10 thermal infrared bands (7-16) from GOES-16 and GOES-18 Level-1b products for our dataset. 
Data is retrieved from NOAA's open data program on AWS S3 at \url{https://noaa-goes16.s3.amazonaws.com/index.html#ABI-L1b-RadF/} and 
\url{https://noaa-goes18.s3.amazonaws.com/index.html#ABI-L1b-RadF/}.

\subsubsection{Geo-Kompsat-2A}

South Korea's Multi-Purpose Satellite GEO-KOMPSAT-2A  flying the Advanced Meteorological Imager (AMI) is nearly identical to ABI, in geostationary orbit with 16 bands at a 0.5-2 km spatial resolution \cite{chung2020meteorological}.
Geo-Kompsat-2A provides observations of the eastern hemisphere including Asia and Australia. 
We selected the 10 thermal infrared bands (7-16) from Geo-Kompsat-2A for our dataset. 
Data was collected using the Korean Meteorology Agency's (KMA) API to access the L1B product, \url{http://api.nmsc.kma.go.kr/homepage/html/main/main.do}. 

\subsubsection{SEVIRI}

The Spinning Enhanced Visible and Infrared Imager (SEVIRI) operated by the European Space Agency is in geostationary orbit observing 4 visible and 8 infrared spectral bands \cite{aminou2002msg}.
SEVIRI is an older generation satellite compared to ABI and AMI and provides coverage over Europe and Africa. 
The 8 thermal infrared bands at 3 km resolution are selected for our dataset. 
The EUMETSAT Data Access Client (eumdac) was used through the python API to access SEVIRI data: \url{https://user.eumetsat.int/resources/user-guides/eumetsat-data-access-client-eumdac-guide}.

\subsubsection{VIIRS}

The Visible Infrared Imaging Radiometer Suite (VIIRS) collects visible, near-infrared, and infrared imagery flying on NOAA's JPSS and Suomi-NPP low earth orbit platforms. 
In polar orbit, VIIRS collects information in areas where geostationary sensors are not available. 
VIIRS L1b on NOAA-20 (VNP02MOD) processed by NASA was collected for this study:   \url{https://ladsweb.modaps.eosdis.nasa.gov/archive/allData/5200/VNP02MOD}. 
Data granules available every 6 minutes are downloaded and reprojected from 750 meters to 16 kms.
Bands M08, M10, M11, M12, M13, M14, M15, and M16 are selected for \ourmodel.

\subsubsection{ATMS}

The Advanced Technology Microwave Sounder (ATMS) provides 3D atmospheric information in all-sky conditions \cite{weng2013calibration}. 
ATMS is onboard the NOAA-20 and Suomi-NPP platforms and provides critical soundings to operational forecasts. 
Level 1b data was collected from NASA in 6-minute granules at \url{https://sounder.gesdisc.eosdis.nasa.gov/data/SNPP_Sounder_Level1/SNPPATMSL1B.3/} and
\url{https://sounder.gesdisc.eosdis.nasa.gov/data/JPSS1_Sounder_Level1/SNDRJ1ATMSL1B.3/}.
Data from both satellites are placed on the same common grid and averaged where duplicate values are observed. 
Limb correction is applied to each swath during preprocessing. 
This generates a low earth orbit modality with more complete data.

\subsubsection{MiRS}

The Microwave integrated Retrieval System (MiRS) is an algorithm applicable to general atmospheric sounding imagery. 
MiRS uses radiative transfer modeling and 1D variational assimilation to translate level 1 brightness temperatures to geophysical variable representing the enviroment.
MiRS produces variables including 3D atmospheric temperature, humidity, dew point, and snow/ice.
It also generates the 2D fields of temperature, total precipitable water, cloud liquid water, and more.
Data was collected from NOAA's Big Data Program on AWS S3 at \url{https://noaa-nesdis-n20-pds.s3.amazonaws.com/index.html#NPR_MIRS_SND/}. 
September 2023 through March 2024 was available for this study.

\subsection{Verification data}

\subsubsection{Radiosondes}
Radiosondes are balloon-bourne instruments that produce direct measurements of atmospheric conditions including temperature, humidity, and wind speed and direction as they ascend through the atmosphere.
The Integrated Global Radiosonde Archive (IGRA) collects and standardizes radiosonde observations from more than 2,800 globally distributed stations.
Data was collected from IGRA at \url{https://www.ncei.noaa.gov/data/integrated-global-radiosonde-archive/access/data-y2d/} for the two month validation period of February to March 2024.

% \subsubsection{Aircraft}

\section{Data Assimilation}

Here we provide a mathematical formulation of DA and view it in a Bayesian perspective. \cite{carrassi2018data} detailed overview of DA is used as reference and recommended to the reader for more information. 

\subsection{Traditional Data Assimilation}

DA, also known more generally as state-estimation theory, uses observations $y$ to optimize the state  $x$ of a dynamical model. This process requires translating noisy observations to the model state while modeling the data with minimal uncertainty.
In NWP, the dynamical model $\mathcal{M}$ is a set of Navier-Stokes equations defining the Earth system. In this overview it is sufficient to define the dynamical model as a general process:
\begin{equation}
    x_k = \mathcal{M}_{k:k-1}(x_{k-1}, \lambda) + \eta_k
\end{equation}
\noindent Here $\lambda$ are parameter vectors that are defined and typically non-trainable parameters. 
$\eta_k$ are model errors representing the error between the model state and true initial conditions.
$x(t_k) = x_k$ are the model state at discrete times.
The observation vector $y_k$ can be split into signal and noise:
\begin{equation}
    y_k = \mathcal{H}_k(x_k) + \epsilon_k
\end{equation}
\noindent Observational operator $\mathcal{H}$ translates the model state to observations.
For satellite imagery, $\mathcal{H}$ is typically a radiative transfer model mapping infrared/microwave channels to geophysical variables like humidity and temperature. 
In current methods, parameters of $\mathcal{H}$ are not trainable and tuned by science teams.

Our approach inherently includes trainable parameters for $\mathcal{H}$ encoding directly from level 1 satellite observations.
$\mathcal{M}$ is learned with the foundation model with the capability of performing forecasting and interpolation.

\subsection{Bayesian DA}

Following \cite{carrassi2018data}, the Bayesian formulation for the unknown process $x$ on the data $y$ can be written as:
\begin{equation}
    p(x|y) = \dfrac{p(y|x) p(x)}{p(y)}
\end{equation}
\noindent Here $p(x)$ is the prior, gathers information from historical observations, typically using physics based principles. 
$p(y|x)$ is the likelihood of the data given the system's state $x$. 
$p(y)$ is the marginal distribution of the observations, $p(y) = \int dx p(y|x)p(x)$. 
Traditionally, it is assumed that $x$ and $y$ are independent and therefore $p(y)$ is treated as a normalization coefficient.

\subsection{Variational DA}

3D/4D variational methods are used in NWP to maximize $p(x|y)$, following \cite{carrassi2018data}. 
The variational problem is defined as:
\begin{equation}
    x_{K:0}^a = \text{argmin}\big(\mathcal{J}(x_{K:0})\big)
\end{equation}
within the observing window $[t_0,t_K]$ and where the cost function $\mathcal{J}(x_{K:0})$ minimizes the model and observation errors.
The cost function is written as:
\begin{equation}
    \mathcal{J}(x_{K:0}) = \frac{1}{2} \sum_{k=0}^K || y_k - \mathcal{H}(x_k)||_{R_k^{-1}}^2 + \frac{1}{2} \sum_{k=0}^K || x_k - \mathcal{M}(x_k)||_{Q_k^{-1}}^2 + \frac{1}{2} || x_0 - x^b ||_{B^{-1}}^2
\end{equation}
\noindent where $R_k$ and $Q_k$ are observational and model errors, assumed to be uncorrelated in time and Gaussian distributed. 
With a long derivation, one finds that minimizing this function is non-trivial bottlenecked by computing large covariance matrices.

\subsection{Data Assimilation and Masked Autoencoders}

In the Bayesian DA approach, it is assumed that the model and observations are independent and as both $\mathcal{M}$ and $\mathcal{H}$ are ridged functions. 
Rather than making this assumption, we can aim to maximize the joint probability of the model and observations, $p(x,y)$. 
Further, let us assume that the model state $x$ consists of a set of geophysical data available for training. 
In this case, our goal is to maximize the probablity of all the data occurring within the same model $\mathcal{M}$ and observational operators $\mathcal{H}$.

Masked pre-training has been found to learn generalizable representations of data across domains.
The masked pre-training objective is trained by randomly removing data and learning to reconstruct the missing values \cite{germain2015made}.
This can be written probabilistically as a a product of nested conditionals:
\begin{equation}
    p(z) = \prod_{d=1}^D p(z_d | z_{\notin d})
\end{equation}
\noindent where $z = [x, y]$ and $z_d$ is a random set of features. 
It has been shown theoretically that the cumulative expected loss of masked pre-training is equivalent to the log-marginal likelihood of the model when using self-predictive conditional probabilities \cite{moreno2023masked}. 
As noted by \cite{moreno2023masked}, variational inference removes the marginal likelihood out of convenience, as seen in variational DA, but leads to sub-optimal optimization.
We show in this work that multi-modal masked autoencoders, including both observations and geophysical variables as data, learns to maximize the marginal likelihood with fewer assumptions that variational DA.

\section{\ourmodel}

This section details the model architecture of \ourmodel. 
The model is a multi-modal masked autoencoder capable of ingesting general Earth data for an any-to-any mapping.
The model was trained on $M=9$ modalities such that each modality is denoted as $X_m$.
3D modalities have time, space, and channel dimensions while the 2D modality includes space and channel dimensions.
Due to memory constraints, an image size of 144 by 144 is used as the model input size.
Each modality is normalized to a Gaussian distribution with $\mu=0$ and $\sigma=1$.

\subsection{Encoding}

The multi-modal data detailed in \ref{appendix:data} consists of high-dimensional 2D/3D modalities. 
Encoder and decoders denoted as $E_m(X_m)$ and $D_m(X_m)$ are defined to create a common token size of 768 across modalities.
Each modality is encoded with a vision transformer and pre-trained using a variational autoencoder (VAE). 
VAEs are trained with a Gaussian latent distribution and $kl\_weight$=$1e-4$. 

\noindent 3D input modalities are encoded using a spatio-temporal vision transformer with token size ($C_m$, 1, 16, 16) and a sin-cosine embedding. 
For these modalities, $E_m(X_m)$ consists of a 3D convolution tokenizer followed by a transformer block with 6 heads.
2D input modalities include space and channel dimensions and encoded using a vanilla vision transformer with token size ($C_m$, 16, 16).
Similarly, for these modalities, $E_m(X_m)$ consists of a 2D convolution tokenizer followed by a transformer block with 6 heads.

\subsection{Multi-modal Transformer}

Once encoded, tokens are concatenated along with a global embedding and fed through a common transformer model.
The transformer consists of 12 blocks with each having $num\_heads$=12. 
Any tokens that have missing data (NaN values) are masked as zeros and therefore ignored by the transformer encoder.
During the training process, a subset of tokens are randomly removed using the same process. 

\subsection{Decoding}

Outputs from the common transformer encoder are used to decode each modality.
For a given modality $M$, all other modalities are used as contextual information.
Context tokens are embedded with dimension 384 and an embedding parameter per decoder which is used to exchange information across sensors.  
6 decoder transformer blocks with $num\_heads$=8 translate the contextual tokens to a completely gap-filled predictions.

\section{Training details}

\ourmodel is trained in stages to maximize the amount of available training data. 
First, encoding and decoding VAEs are pretrained on all modalities for 12-24 hours on single GPUs. 
Next, all data outside of MiRS is available for all of 2022 and 2023 for training, including all level 1 remote sensing observations.
This data is used to train the base foundation model for 2 weeks on 16 GPUs.
Lastly, MiRS modalities are added from September 2023 to January 2024 for further training, still using a masked autoencoder.
The output of this training process is then used for evaluation and data assimilation.

\subsection{Masking}

It has been found that a substantial percentage of tokens can be removed while training masked autoencoders.
In the spatio-temporal context, it has been found that between 95\% and 99\% of tokens may be removed to learn expressive representations with low reconstruction error. 
MMAE and \ourmodel takes advantage of this property by dropping tokens from each modality with random ratios using Dirichlet sampling. 

With a token size of (1, 16, 16), each 3D modality has a total of 976 tokens. 
All together, this adds up to 8,752 tokens for every training sample. 
Approximately 20\% of the tokens have complete data which provides 1,750 useful tokens per sample. 
During training we select 128 of these tokens to reproduce the remainder of the data, dropping 98.6\% of the data.
Masking is performed by sampling $K_m$ tokens per modality, such that $K_m$ is sampled from a Dirichlet distribution. 
As the results show, the model is capable of reconstructing all the data from just 1.4\% of an input sample.

\subsection{Loss and optimization}

A mean square error loss is applied to masked tokens that do not including missing data.

\begin{equation}
    \label{eq:loss}
    \text{L}(\Theta) = \frac{1}{N} \sum_m (X_m - \hat{X}_m)^2
\end{equation}

\noindent Adam optimization was used for training \ourmodel \cite{kingma2014adam} with parameters $\beta_1=0.5$ and $\beta_2=0.9$. 

\section{Verification}

\subsection{Background departure statistics}
\label{sup:sec:backgrounddepartures}

Evaluate \ourmodel's background departures by performing a 1 hour forecast and comparing with observations. 
Predictions were made for 0 and 1 hour forecasts, corresponding to analysis and background.
These are each compared with observations in the test set.
Statistics of bias and mean absolute error are presented here for every channel and pressure level. 
Tables \ref{tab:leo_stats}, \ref{tab:geo_stats},  \ref{tab:mirs_temp_stats}, and \ref{tab:mirs_vapor_stats} present these statistics for LEO, GEO, temperature and specific humidity, respectively.

\subsection{Global reanalysis}
\label{sup:global_reanalysis}

We performed model verification by comparison to held out (from \ourmodel) radiosonde observations. 
Temperature and humidity data from ERA5, MERRA-2, MiRS, \ourmodel, and radiosondes were collected on a 0.16$^\circ$ global grid and compared for the test period comprising of February and March 2024.
Relative humidity was calculated for \ourmodel and MiRS using the World Meteorological Organization definition based on vapor partial pressures.
Unless otherwise noted, plots and error statistics are based on the subset of simulations and observations that coincide in all 5 datasets.
Table \ref{tab:radiosonde_results} presents summary statistics for temperature and relative humidity errors compared to radiosondes.

\begin{figure}[h!]
    \centering
    \includegraphics[width=\linewidth]{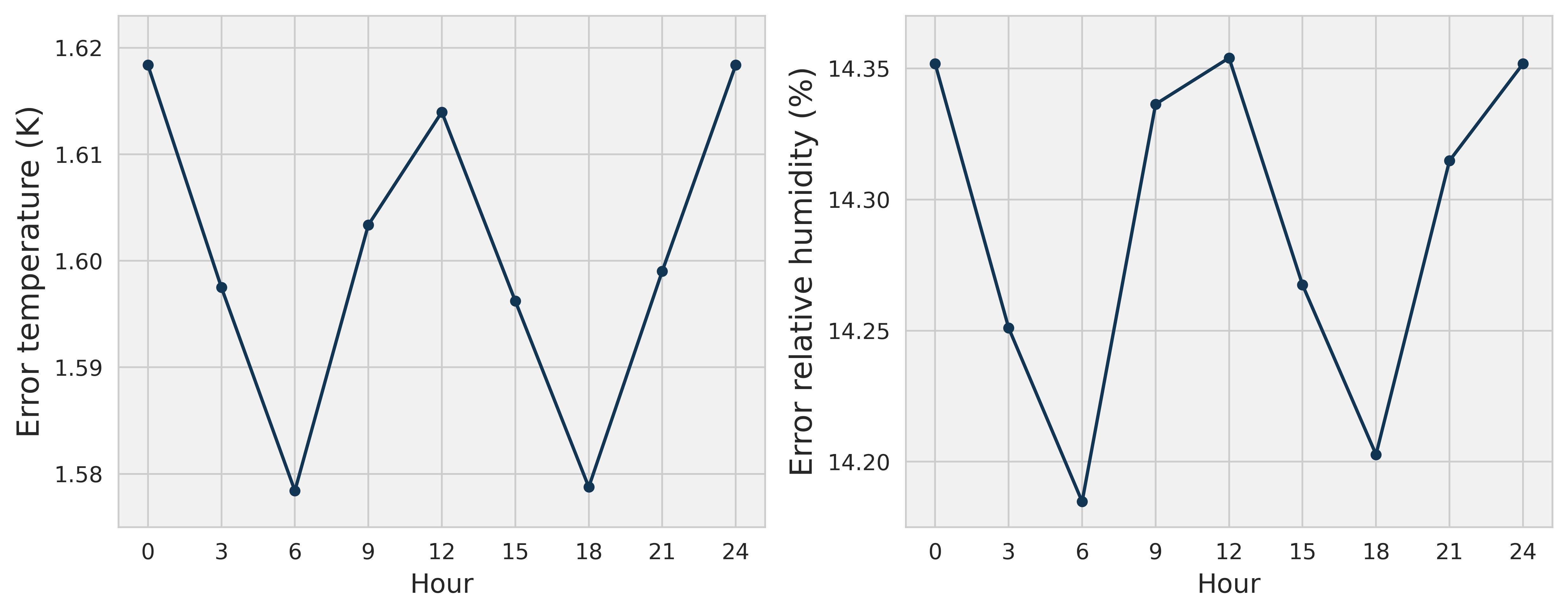}
    \caption{\small{\textbf{Analysis of \ourmodel's hourly performance} shows minimized error during the middle of the non-overlapping 12-hour periods used for inference.}}
    \label{fig:hourly}
\end{figure}

Inference was applied on 12 hour non-overlapping periods over the test period of February and March 2024. 
The time blocks used for inference were [00:00, 12:00) and [12:00, 24:00) GMT.
Analysis of the average hourly error between \ourmodel and MERRA-2 temperature and humidity shows that \ourmodel error 
is minimized mid-period and peaks at the beginning and end of each 12-hour block (Figure \ref{fig:hourly}).
This corresponds to the maximum amount of temporal context available at hour 06:00 (or 18:00).
This effect could be ameliorated by a more computationally intensive inference approach, such as a moving window approach that discards the initial and final time frames.
However, the effect is relatively small in size accounting for $\sim$2.5\% of temperature error and 1.2\% of relative humidity error as compared to MERRA-2.

\begin{figure}[h!]
    \centering
    \includegraphics[width=0.47\linewidth]{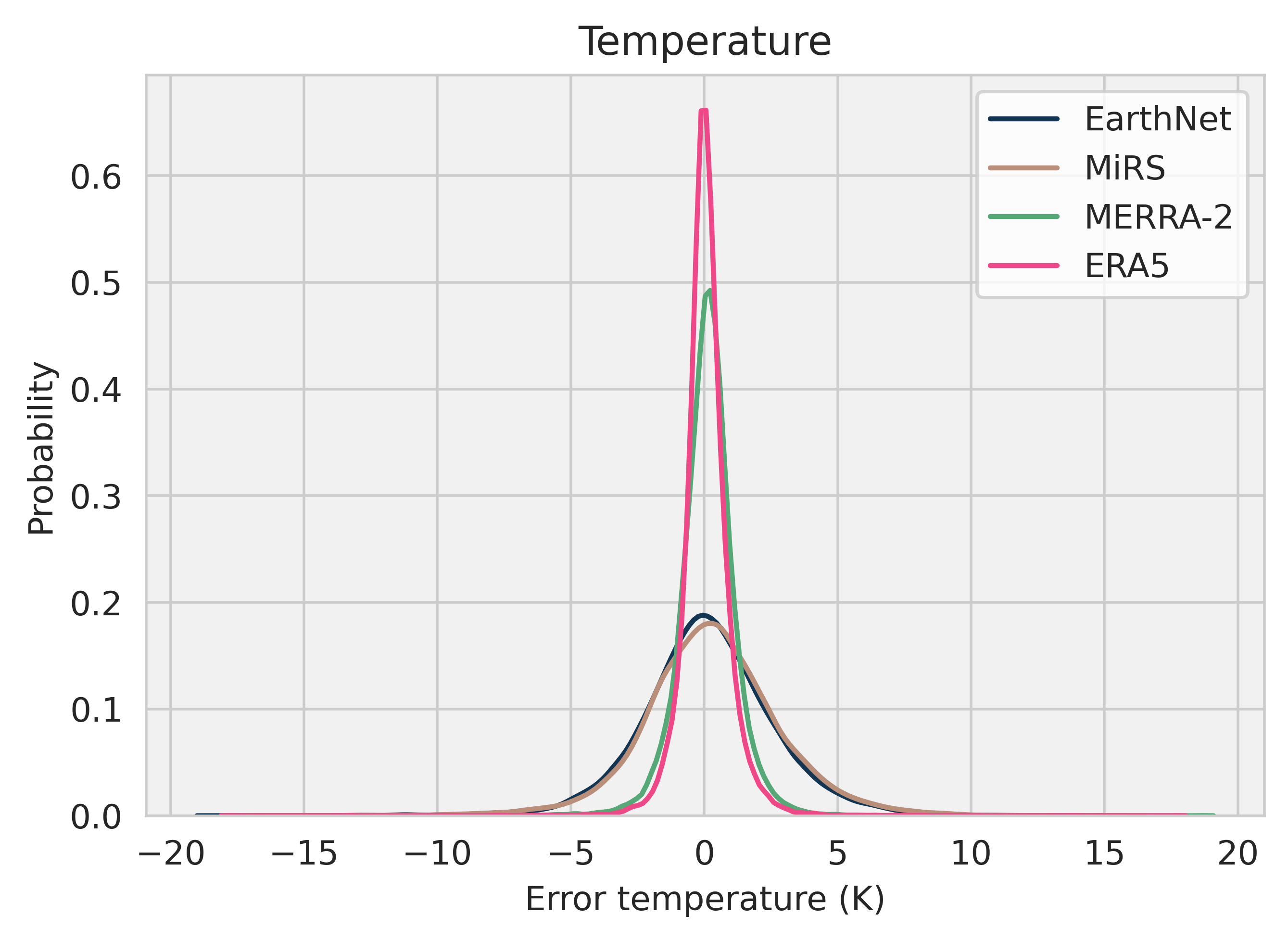}
    \includegraphics[width=0.48\linewidth]{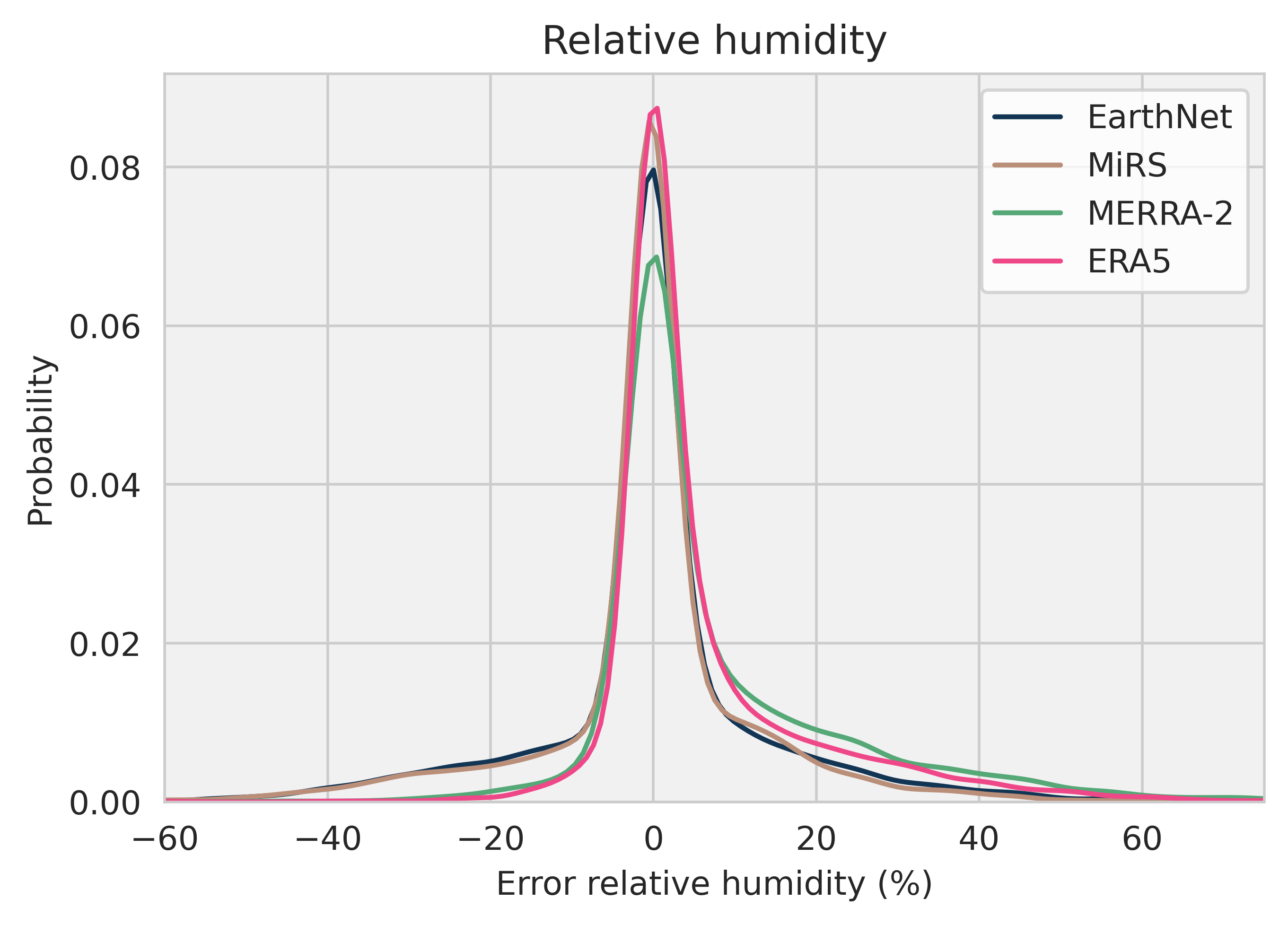}
    \caption{\small{\textbf{Distributions of errors} (analysis minus radiosonde observations) show that temperature error distributions are symmetrical but wider for \ourmodel and MiRS than the NWP models.
    Relative humidity errors have more similar spread among the evaluated models.}}
    \label{fig:error_probability}
\end{figure}

We further analyzed the probability distribution of errors (analysis - radiosonde observations) for each of EarthNet, MiRS, ERA5 and MERRA-2 (Figure \ref{fig:error_probability}).
We found that for temperature, error distributions are roughly symmetrical for each data source, with MERRA-2/ERA5 having a smaller spread of errors than \ourmodel/MiRS, as also indicated by the scatter plots in Figure \ref{fig:reanalysis}c.
For relative humidity, we observe that \ourmodel and MiRS have more underestimation errors, while ERA5/MERRA-2 have more overestimation errors.

\begin{figure}[h!]
    \centering
    \includegraphics[width=\linewidth]{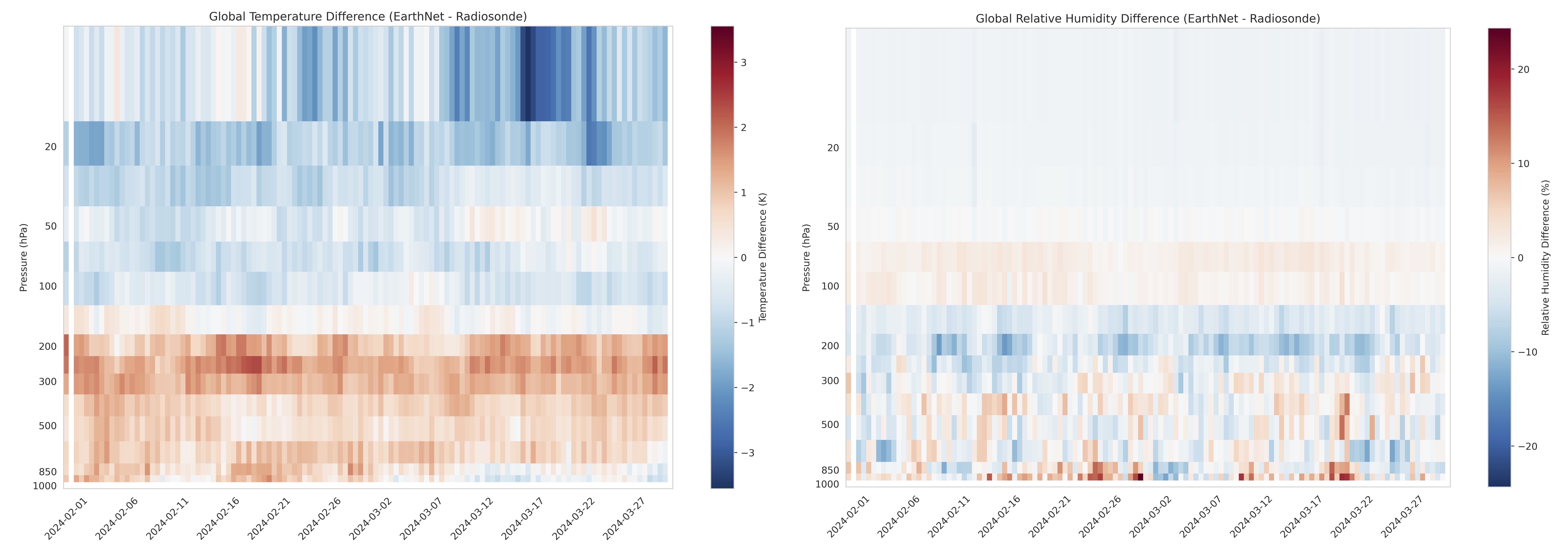}
    \caption{\small{\textbf{Global bias errors by time and level} (\ourmodel minus radiosonde observations). }}
    \label{fig:error_time_lev}
\end{figure}

\begin{figure}[h!]
    \centering
    \includegraphics[width=\linewidth]{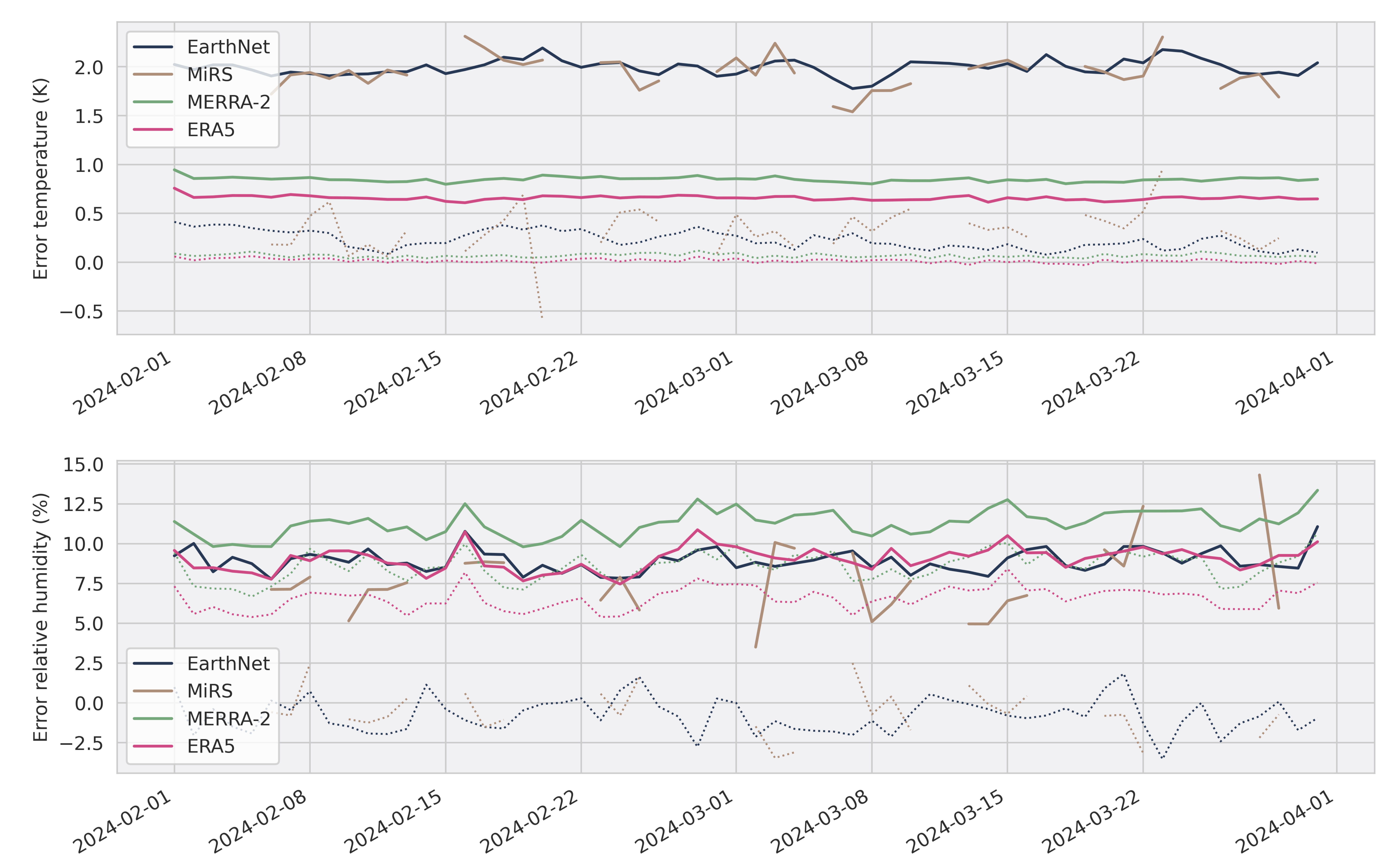}
    \caption{\small{\textbf{Time series} of daily mean absolute error (thick lines) and daily mean bias error (thin dotted lines) compared to radiosonde observations.}}
    \label{fig:time_series}
\end{figure}

To assess the temporal patterns of error by level of the atmosphere, we plotted time series of mean bias error of \ourmodel compared to radiosonde observations in Figure \ref{fig:error_time_lev}.
We see that the direction of the temperature bias is consistently negative at 100 hPa and above and consistently positive between 200 and 850 hPa.
In contrast, relative humidity bias errors do not follow a consistent direction between 300 hPa and the surface, but are alternately positive and negative over time.
Finally, time series of global error (mean absolute and mean bias) show that the errors compared to radiosondes are relatively stable over time (Figure \ref{fig:time_series}). 
For the analysis in Figure \ref{fig:time_series} we used all data from each source that coincided with radiosonde observations.

% \section{Data Availability}

% \section{Code availability}

% \section{Visualizations}

% Jupyter notebook figures - time series, vertical profiles, soundings for extreme events

% https://www.weather.gov/source/zhu/ZHU_Training_Page/convective_parameters/skewt/skewtinfo.html#:~:text=Snow%20Sounding,do%20not%20rise%20from%20surface.
\subsection{Atmospheric Soundings}

\begin{figure}[h!]
    \centering
     \begin{subfigure}[b]{0.49\textwidth}
         \centering
         \includegraphics[width=\linewidth]{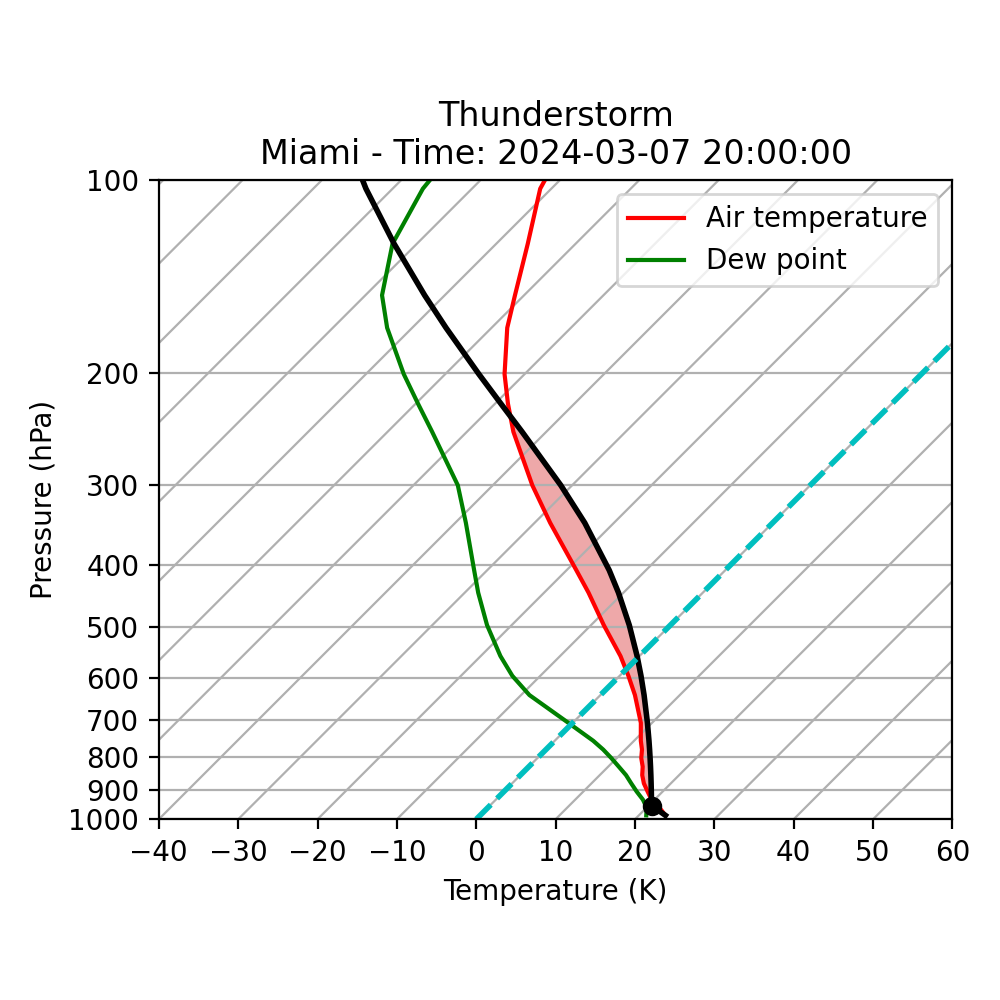}
         \caption{A thunderstorm in Miami, FL (CAPE=1500 j/kg).}
         \label{fig:sounder:miami}
     \end{subfigure}
     \hfill
     \begin{subfigure}[b]{0.49\textwidth}
         \centering
         \includegraphics[width=\linewidth]{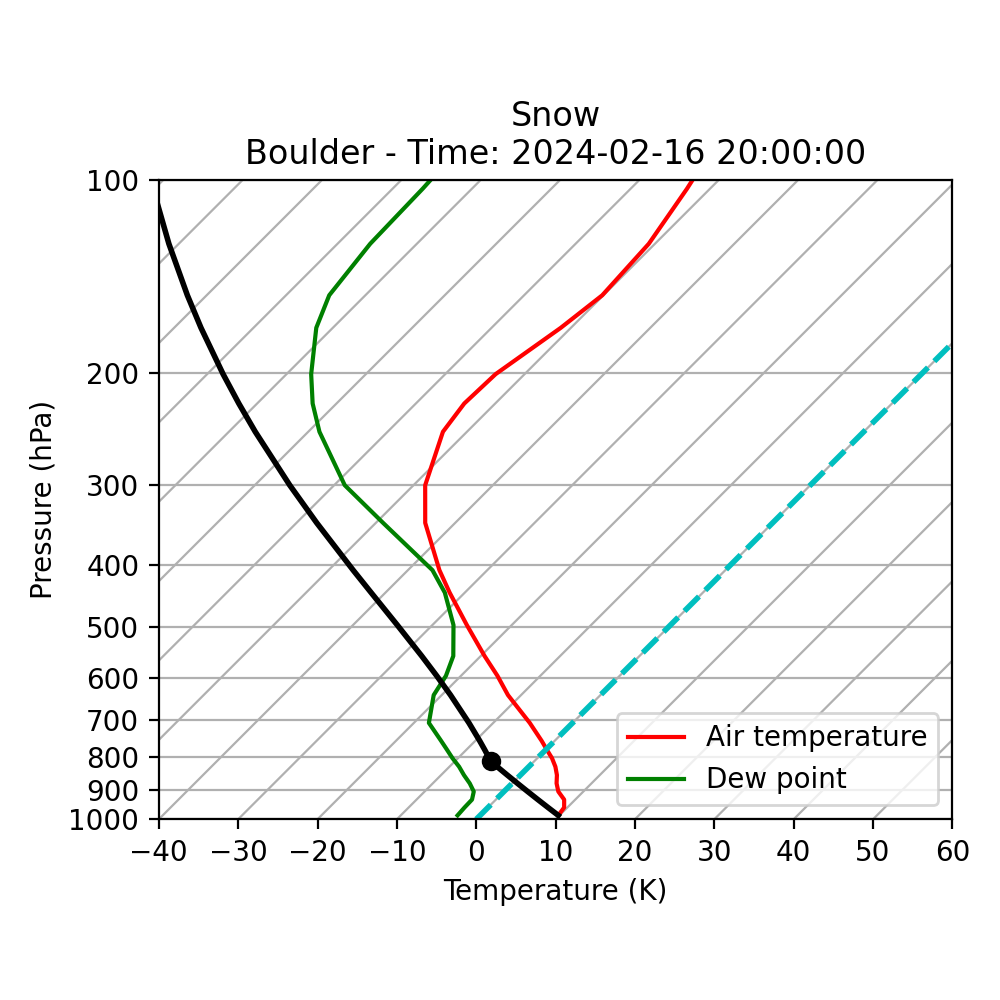}
         \caption{A snow event in Boulder, Colorado.}
         \label{fig:sounder:boulder}
     \end{subfigure}
     \vfill
     \begin{subfigure}[b]{0.49\textwidth}
         \centering
         \includegraphics[width=\linewidth]{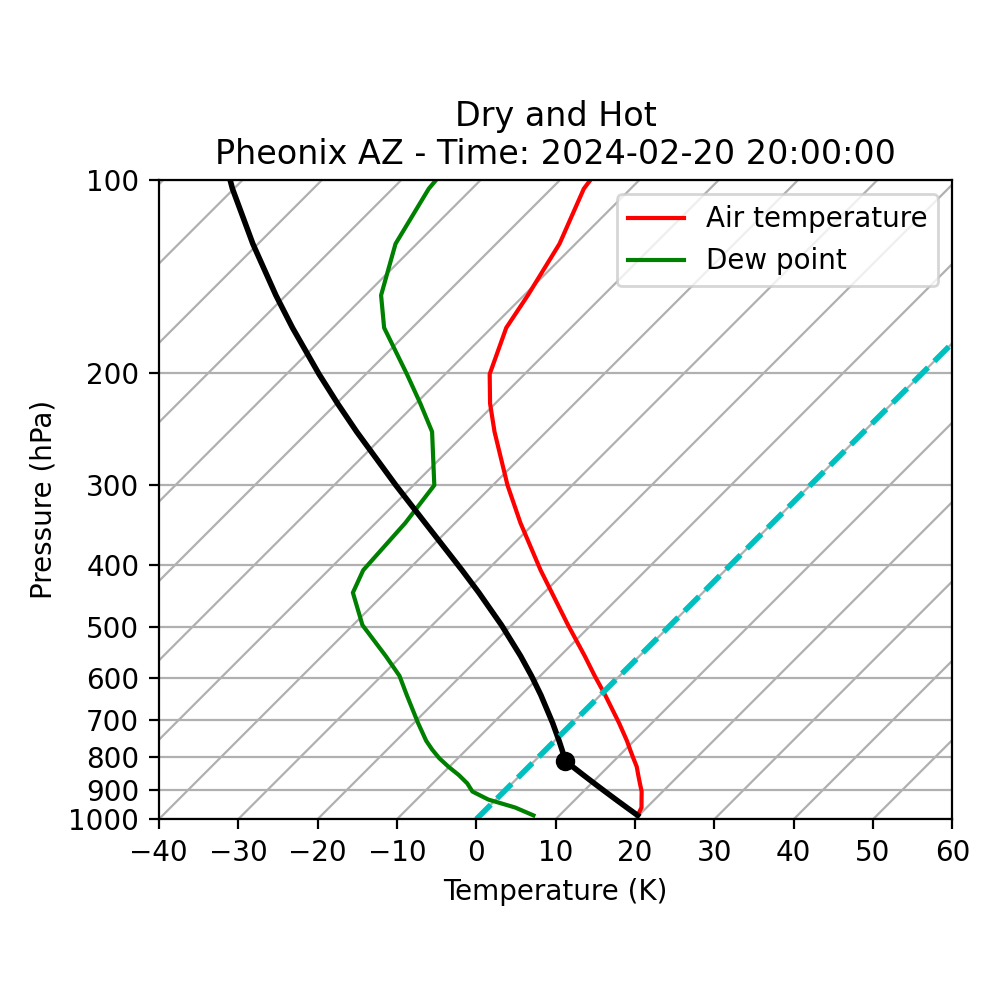}
         \caption{A dry day in Pheonix, Arizona.}
         \label{fig:sounder:pheonix}
     \end{subfigure}
     \hfill
     \begin{subfigure}[b]{0.49\textwidth}
         \centering
         \includegraphics[width=\linewidth]{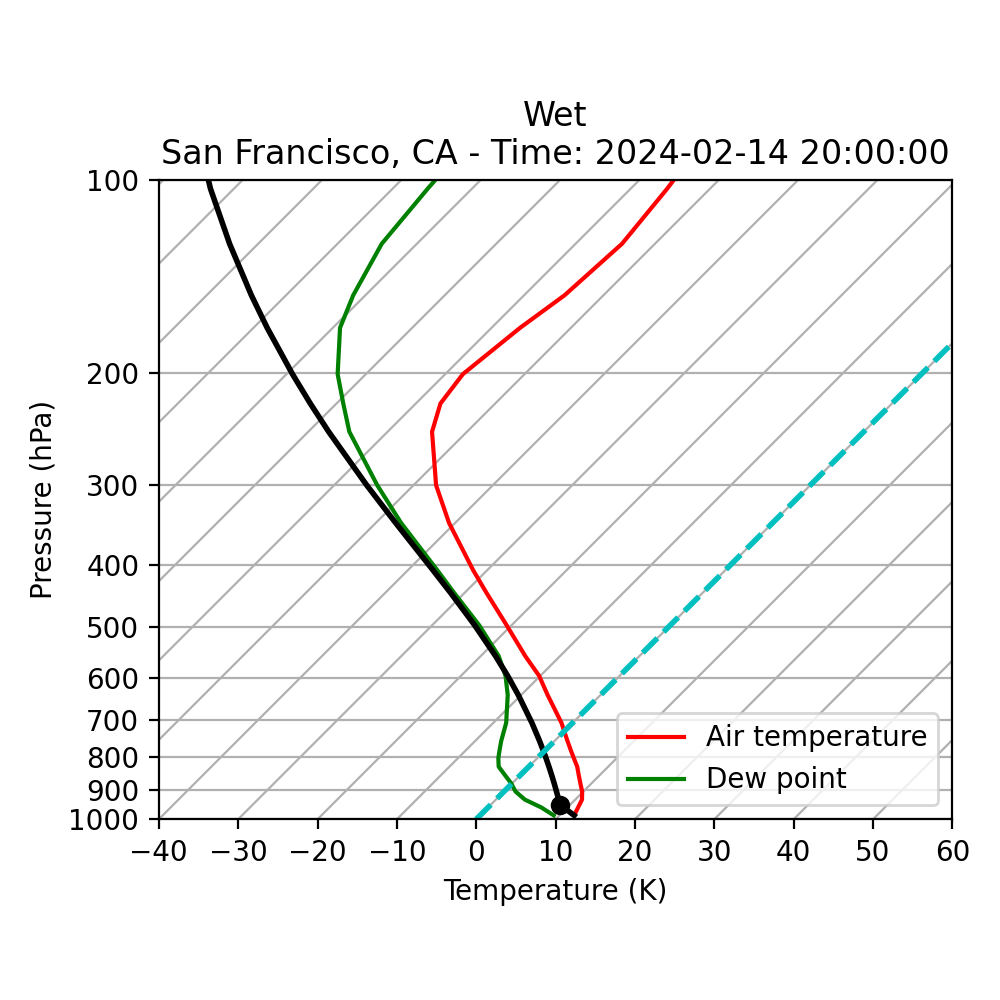}
         \caption{A wet day in San Fransisco, California.}
         \label{fig:sounder:sf}
     \end{subfigure}
    \caption{\small{\textbf{Soundings from \ourmodel} of four atmospheric events produced using metpy.}}
    \label{fig:soundings}
\end{figure}

Temperature and specific humidity profiles can be converted to dew point temperature and used to analyze weather conditions.
Figure \ref{fig:soundings} shows atmospheric soundings of four weather events using the metpy python package.
Air temperature and dew point temperature are plotting with the parcel lapse rate.

\section{Tables}

\begin{table}[h!]
    \centering
\begin{tabular}{llcccc}
\toprule
\multirow{2}{*}{\textbf{Sensor}} & 
\multirow{2}{*}{\textbf{Channel}} & 
\multicolumn{2}{c}{\textbf{Bias (K)}} & 
\multicolumn{2}{c}{\textbf{Mean Absolute Error (K)}} \\
\cmidrule(lr){3-4} \cmidrule(lr){5-6}
& & {\textbf{EarthNet}} & {\textbf{Background}} & {\textbf{EarthNet}} & {\textbf{Background}} \\
\midrule
ATMS & 1  & 0.373 & 0.352 & 3.082 & 3.793 \\
 & 2  & 0.395 & 0.313 & 3.094 & 3.861 \\
 & 3  & 0.105 & -0.026 & 2.055 & 2.396 \\
 & 4  & 0.099 & 0.006 & 1.401 & 1.654 \\
 & 5  & 0.096 & 0.043 & 0.894 & 1.240 \\
 & 6  & -0.038 & -0.035 & 0.750 & 1.268 \\
 & 7  & -0.004 & -0.000 & 0.351 & 0.702 \\
 & 8  & -0.014 & -0.001 & 0.254 & 0.358 \\
 & 9  & -0.005 & -0.004 & 0.359 & 0.611 \\
 & 10  & 0.010 & -0.006 & 0.471 & 0.751 \\
 & 11  & 0.042 & 0.014 & 0.441 & 0.848 \\
 & 12  & 0.039 & 0.001 & 0.518 & 1.203 \\
 & 13  & 0.014 & -0.034 & 0.740 & 1.617 \\
 & 14  & -0.062 & -0.126 & 1.045 & 2.016 \\
 & 15  & -0.024 & -0.102 & 1.169 & 1.316 \\
 & 16  & -0.044 & 0.038 & 3.062 & 3.531 \\
 & 17  & 0.134 & -0.230 & 3.809 & 4.418 \\
 & 18  & -0.010 & -0.159 & 2.436 & 2.871 \\
 & 19  & 0.048 & -0.022 & 1.970 & 2.236 \\
 & 20  & -0.034 & -0.140 & 1.532 & 1.882 \\
 & 21  & -0.011 & -0.085 & 1.167 & 1.383 \\
 & 22  & -0.051 & -0.059 & 1.029 & 1.132 \\
 \midrule
VIIRS & M08 & -0.296 & -0.370 & 1.792 & 1.957 \\
 & M10 & -0.220 & -0.216 & 1.393 & 1.468 \\
 & M11 & -0.232 & -0.245 & 1.252 & 1.327 \\
 & M12 & -0.430 & -0.441 & 4.840 & 5.023 \\
 & M13 & -0.225 & -0.509 & 3.203 & 3.365 \\
 & M14 & -0.009 & -0.300 & 3.344 & 3.477 \\
 & M15 & -0.037 & -0.246 & 3.631 & 3.750 \\
 & M16 & -0.037 & -0.230 & 3.684 & 3.799 \\
\bottomrule
\end{tabular}
    \caption{LEO background error statistics.}
    \label{tab:leo_stats}
\end{table}

\begin{table}[h!]
    \centering
\begin{tabular}{llcccc}
\toprule
\multirow{2}{*}{\textbf{Sensor}} & 
\multirow{2}{*}{\textbf{Channel}} & 
\multicolumn{2}{c}{\textbf{Bias (K)}} & 
\multicolumn{2}{c}{\textbf{Mean Absolute Error (K)}} \\
\cmidrule(lr){3-4} \cmidrule(lr){5-6}
& & {\textbf{EarthNet}} & {\textbf{Background}} & {\textbf{EarthNet}} & {\textbf{Background}} \\
\midrule
GOES-16 & 7  & -0.200 & -0.235 & 4.918 & 5.641 \\
 & 8  & 0.029 & 0.051 & 1.541 & 1.643 \\
 & 9  & 0.063 & 0.093 & 2.035 & 2.180 \\
 & 10  & 0.062 & 0.101 & 2.547 & 2.723 \\
 & 11  & 0.107 & 0.136 & 4.973 & 5.301 \\
 & 12  & 0.029 & 0.047 & 2.809 & 2.967 \\
 & 13  & 0.134 & 0.177 & 5.270 & 5.621 \\
 & 14  & 0.168 & 0.214 & 5.367 & 5.727 \\
 & 15  & 0.154 & 0.190 & 5.180 & 5.531 \\
 & 16  & 0.024 & 0.056 & 3.703 & 3.947 \\
 \midrule
 GOES-18 & 7  & -0.050 & -0.281 & 4.387 & 4.969 \\
 & 8  & 0.033 & 0.040 & 1.442 & 1.524 \\
 & 9  & 0.027 & 0.032 & 1.893 & 2.014 \\
 & 10  & 0.073 & 0.084 & 2.303 & 2.445 \\
 & 11  & 0.117 & 0.117 & 4.605 & 4.863 \\
 & 12  & -0.003 & -0.022 & 2.684 & 2.811 \\
 & 13  & 0.121 & 0.131 & 4.879 & 5.156 \\
 & 14  & 0.131 & 0.149 & 4.934 & 5.223 \\
 & 15  & 0.117 & 0.132 & 4.742 & 5.023 \\
 & 16  & 0.116 & 0.129 & 3.375 & 3.570 \\
\midrule
SEVIRI & IR\_016  & -0.223 & -0.419 & 2.426 & 3.375 \\
 & IR\_039  & -0.333 & -0.403 & 4.945 & 5.758 \\
 & IR\_087  & -0.224 & -0.258 & 5.699 & 6.113 \\
 & IR\_097  & -0.169 & -0.225 & 3.199 & 3.424 \\
 & IR\_108  & -0.206 & -0.245 & 6.051 & 6.520 \\
 & IR\_120  & -0.204 & -0.245 & 6.031 & 6.504 \\
 & IR\_134  & -0.154 & -0.175 & 3.752 & 4.023 \\
 & WV\_062  & -0.118 & -0.136 & 1.699 & 1.812 \\
 & WV\_073  & -0.107 & -0.107 & 2.859 & 3.059 \\
 \midrule
GK2A & 7  & -0.086 & -0.287 & 4.590 & 5.234 \\
 & 8  & 0.076 & 0.095 & 1.541 & 1.633 \\
 & 9  & 0.102 & 0.102 & 2.041 & 2.184 \\
 & 10  & 0.118 & 0.113 & 2.518 & 2.691 \\
 & 11  & 0.002 & -0.015 & 4.883 & 5.219 \\
 & 12  & 0.084 & 0.086 & 2.820 & 2.980 \\
 & 13  & 0.010 & 0.004 & 5.176 & 5.543 \\
 & 14  & -0.020 & -0.020 & 5.258 & 5.629 \\
 & 15  & 0.004 & -0.001 & 5.012 & 5.359 \\
\bottomrule
\end{tabular}
     \caption{GEO background error statistics.}
    \label{tab:geo_stats}
\end{table}

\begin{table}[h!]
    \centering

\begin{tabular}{llcccc}
\toprule
\multirow{2}{*}{\textbf{Variable}} & 
\multirow{2}{*}{\textbf{Pressure}} & 
\multicolumn{2}{c}{\textbf{Bias (K)}} & 
\multicolumn{2}{c}{\textbf{Mean Absolute Error (K)}} \\
\cmidrule(lr){3-4} \cmidrule(lr){5-6}
& & {\textbf{EarthNet}} & {\textbf{Background}} & {\textbf{EarthNet}} & {\textbf{Background}} \\
\midrule
Temperature & 1.000 & -0.162 & -0.191 & 2.556 & 2.581 \\
 & 2.200 & 0.361 & 0.400 & 1.756 & 1.786 \\
 & 2.700 & 0.278 & 0.313 & 1.622 & 1.649 \\
 & 4.900 & -0.170 & -0.187 & 1.617 & 1.646 \\
 & 7.000 & -0.212 & -0.236 & 1.487 & 1.519 \\
 & 9.500 & -0.147 & -0.174 & 1.078 & 1.097 \\
 & 20.900 & -0.083 & -0.110 & 0.744 & 0.752 \\
 & 29.100 & 0.024 & 0.004 & 0.674 & 0.675 \\
 & 51.500 & -0.227 & -0.254 & 0.727 & 0.728 \\
 & 71.500 & -0.204 & -0.227 & 0.751 & 0.751 \\
 & 103.000 & 0.049 & 0.043 & 0.635 & 0.631 \\
 & 125.600 & 0.025 & 0.011 & 0.667 & 0.671 \\
 & 151.300 & -0.034 & -0.051 & 0.713 & 0.721 \\
 & 170.100 & -0.101 & -0.123 & 0.746 & 0.756 \\
 & 201.000 & -0.303 & -0.335 & 0.863 & 0.878 \\
 & 223.400 & -0.441 & -0.472 & 0.975 & 0.995 \\
 & 247.400 & -0.363 & -0.384 & 1.012 & 1.030 \\
 & 300.000 & -0.034 & -0.030 & 0.790 & 0.795 \\
 & 343.600 & -0.018 & -0.004 & 0.725 & 0.725 \\
 & 407.500 & -0.173 & -0.148 & 0.838 & 0.835 \\
 & 441.900 & -0.172 & -0.142 & 0.878 & 0.874 \\
 & 496.600 & -0.109 & -0.072 & 0.879 & 0.876 \\
 & 555.200 & -0.008 & 0.033 & 0.880 & 0.883 \\
 & 596.300 & 0.057 & 0.100 & 0.907 & 0.914 \\
 & 639.100 & 0.296 & 0.338 & 1.076 & 1.091 \\
 & 706.600 & 0.456 & 0.504 & 1.242 & 1.263 \\
 & 753.600 & 0.508 & 0.559 & 1.329 & 1.357 \\
 & 777.800 & 0.523 & 0.574 & 1.377 & 1.408 \\
 & 802.400 & 0.517 & 0.570 & 1.419 & 1.452 \\
 & 827.400 & 0.512 & 0.566 & 1.459 & 1.494 \\
 & 852.800 & 0.509 & 0.562 & 1.511 & 1.546 \\
 & 878.600 & 0.532 & 0.584 & 1.577 & 1.613 \\
 & 904.900 & 0.586 & 0.640 & 1.644 & 1.683 \\
 & 931.500 & 0.682 & 0.744 & 1.714 & 1.760 \\
 & 958.600 & 0.812 & 0.891 & 1.807 & 1.866 \\
 & 986.100 & 0.984 & 1.066 & 1.948 & 2.013 \\
\bottomrule
\end{tabular}
    \caption{MiRS temperature background error statistics.}
    \label{tab:mirs_temp_stats}
\end{table}

\begin{table}[h!]
    \centering
\begin{tabular}{llcccc}
\toprule
\multirow{2}{*}{\textbf{Variable}} & 
\multirow{2}{*}{\textbf{Pressure}} & 
\multicolumn{2}{c}{\textbf{Bias (K)}} & 
\multicolumn{2}{c}{\textbf{Mean Absolute Error (K)}} \\
\cmidrule(lr){3-4} \cmidrule(lr){5-6}
& & {\textbf{EarthNet}} & {\textbf{Background}} & {\textbf{EarthNet}} & {\textbf{Background}} \\
\midrule
Specific Humidity & 1.000 & -0.162 & -0.191 & 2.556 & 2.581 \\
 & 2.200 & 0.361 & 0.400 & 1.756 & 1.786 \\
 & 2.700 & 0.278 & 0.313 & 1.622 & 1.649 \\
 & 4.900 & -0.170 & -0.187 & 1.617 & 1.646 \\
 & 7.000 & -0.212 & -0.236 & 1.487 & 1.519 \\
 & 9.500 & -0.147 & -0.174 & 1.078 & 1.097 \\
 & 20.900 & -0.083 & -0.110 & 0.744 & 0.752 \\
 & 29.100 & 0.024 & 0.004 & 0.674 & 0.675 \\
 & 51.500 & -0.227 & -0.254 & 0.727 & 0.728 \\
 & 71.500 & -0.204 & -0.227 & 0.751 & 0.751 \\
 & 103.000 & 0.049 & 0.043 & 0.635 & 0.631 \\
 & 125.600 & 0.025 & 0.011 & 0.667 & 0.671 \\
 & 151.300 & -0.034 & -0.051 & 0.713 & 0.721 \\
 & 170.100 & -0.101 & -0.123 & 0.746 & 0.756 \\
 & 201.000 & -0.303 & -0.335 & 0.863 & 0.878 \\
 & 223.400 & -0.441 & -0.472 & 0.975 & 0.995 \\
 & 247.400 & -0.363 & -0.384 & 1.012 & 1.030 \\
 & 300.000 & -0.034 & -0.030 & 0.790 & 0.795 \\
 & 343.600 & -0.018 & -0.004 & 0.725 & 0.725 \\
 & 407.500 & -0.173 & -0.148 & 0.838 & 0.835 \\
 & 441.900 & -0.172 & -0.142 & 0.878 & 0.874 \\
 & 496.600 & -0.109 & -0.072 & 0.879 & 0.876 \\
 & 555.200 & -0.008 & 0.033 & 0.880 & 0.883 \\
 & 596.300 & 0.057 & 0.100 & 0.907 & 0.914 \\
 & 639.100 & 0.296 & 0.338 & 1.076 & 1.091 \\
 & 706.600 & 0.456 & 0.504 & 1.242 & 1.263 \\
 & 753.600 & 0.508 & 0.559 & 1.329 & 1.357 \\
 & 777.800 & 0.523 & 0.574 & 1.377 & 1.408 \\
 & 802.400 & 0.517 & 0.570 & 1.419 & 1.452 \\
 & 827.400 & 0.512 & 0.566 & 1.459 & 1.494 \\
 & 852.800 & 0.509 & 0.562 & 1.511 & 1.546 \\
 & 878.600 & 0.532 & 0.584 & 1.577 & 1.613 \\
 & 904.900 & 0.586 & 0.640 & 1.644 & 1.683 \\
 & 931.500 & 0.682 & 0.744 & 1.714 & 1.760 \\
 & 958.600 & 0.812 & 0.891 & 1.807 & 1.866 \\
 & 986.100 & 0.984 & 1.066 & 1.948 & 2.013 \\
\bottomrule
\end{tabular}
    \caption{MiRS specific humidity background error statistics.}
    \label{tab:mirs_vapor_stats}
\end{table}

\begin{table}[htbp]
\centering
\small
\begin{tabular}{
    r
    !{\hspace{1.5em}}
    *{4}{S[table-format=2.2]}
    !{\hspace{1.5em}}
    *{4}{S[table-format=2.2]}
}
\toprule
\multirow{2}{*}{\textbf{Pressure (hPa)}} 
& \multicolumn{4}{c}{\textbf{Temperature Error (K)}} 
& \multicolumn{4}{c}{\textbf{Relative Humidity Error (\%)}} \\
\cmidrule(lr){2-5} \cmidrule(lr){6-9}
& {\textbf{EarthNet}} & {\textbf{MiRS}} & {\textbf{MERRA-2}} & {\textbf{ERA5}} 
& {\textbf{EarthNet}} & {\textbf{MiRS}} & {\textbf{MERRA-2}} & {\textbf{ERA5}} \\
\midrule
925 & 2.38 & 2.61 & 0.87 & 0.62 & 18.58* & 16.91 & 10.39 & 7.59 \\
850 & 2.16 & 2.40 & 0.74 & 0.51 & 19.60* & 17.21 & 9.77 & 7.63 \\
700 & 1.63 & 1.79 & 0.59 & 0.44 & 19.55* & 17.59 & 13.12 & 10.05 \\
500 & 1.56* & 1.58 & 0.59 & 0.40 & 15.17* & 13.21 & 15.97 & 14.69 \\
400 & 1.54* & 1.52 & 0.57 & 0.40 & 16.10* & 14.44 & 20.02 & 19.75 \\
300 & 2.03* & 2.16 & 0.67 & 0.46 & 14.77* & 14.11 & 24.53 & 18.27 \\
250 & 2.61* & 2.80 & 0.82 & 0.54 & 8.04 & 8.49 & 22.03 & 12.13 \\
200 & 2.55* & 2.73 & 0.90 & 0.67 & 4.06* & 4.10 & 12.07 & 7.13 \\
150 & 1.46* & 1.52 & 0.81 & 0.64 & 1.86* & 2.01 & 6.13 & 4.58 \\
100 & 1.55* & 1.71 & 0.68 & 0.61 & 2.47* & 2.30 & 5.96 & 6.00 \\
70 & 1.40 & 1.46 & 0.76 & 0.64 & 2.32* & 2.40 & 5.03 & 4.73 \\
50 & 1.37* & 1.29 & 0.79 & 0.71 & 0.66* & 0.70 & 1.01 & 1.06 \\
30 & 1.82* & 1.61 & 0.92 & 0.81 & 0.55* & 0.54 & 0.34 & 0.34 \\
20 & 1.89* & 1.66 & 1.05 & 0.94 & 0.60* & 0.60 & 0.46 & 0.46 \\
10 & 2.88* & 2.60 & 1.37 & 1.24 & 0.66* & 0.66 & 0.57 & 0.58 \\
\midrule
\textbf{Average} & \textbf{1.92} & \textbf{1.96} & \textbf{0.81} & \textbf{0.64} & \textbf{8.33} & \textbf{7.68} & \textbf{9.83} & \textbf{7.67} \\
\bottomrule
\end{tabular}
\caption{\small{\textbf{3D data verification} for \ourmodel, MiRS, MERRA-2 and ERA5. Temperature and relative humidity are compared to radiosondes at different pressure levels of the atmosphere.  
Starred (*) EarthNet values indicate no statistically significant difference from MiRS errors at the $p<0.05$ level.}}
\label{tab:radiosonde_results}
\end{table}

% \end{bibunit}